\documentclass{article} %
\usepackage{omni-life}
\usepackage{graphicx}
\usepackage{xcolor}
\usepackage{hyperref}
\definecolor{repolink}{HTML}{51469A}
\usepackage{booktabs}
\usepackage{graphicx}
\usepackage{enumitem}
\usepackage{wrapfig}
\usepackage{algorithm}
\usepackage{algpseudocode}
\usepackage{wrapfig}
\usepackage{float}
\usepackage{microtype}
\usepackage{amsmath}
\usepackage{amssymb}
\usepackage{colortbl}
\usepackage[utf8]{inputenc}
\definecolor{lightgray}{rgb}{0.9,0.9,0.9}
\usepackage{caption}
\usepackage{subcaption}
\usepackage{xcolor}
\usepackage{setspace}
\usepackage{url}
\usepackage{multirow}
\usepackage{colortbl}
\usepackage{tabularx}
\usepackage{blindtext}
\usepackage{pgfplots}
\pgfplotsset{compat=1.18} 
\usepackage{tikz}
\usetikzlibrary{er,positioning,bayesnet}
\usepackage{makecell}
\usepackage{tipa}
\usepackage{siunitx}
\usepackage{nicefrac}
\usepackage{tocloft}
\usepackage{listings}
\usepackage[raster,skins]{tcolorbox} %
\usepackage{xltabular}
\usepackage{adjustbox}
\usepackage{xurl}
\definecolor{statSynthetic}{HTML}{B8D9F5}
\definecolor{statRealWorld}{HTML}{C8DDBB}
\definecolor{statTotal}{HTML}{D2C6E5}
\definecolor{statSage}{HTML}{C8D5C0}
\definecolor{statSageLight}{HTML}{E8EDE4}
\definecolor{statTotal}{HTML}{D2C6E5}
\definecolor{statTotalLight}{HTML}{EEE9F5}
\definecolor{ybdeep}{HTML}{12A594}
\definecolor{ybrule}{HTML}{0B7A6B}
\definecolor{ybgreen}{HTML}{0E9E86}
\definecolor{ybtint}{HTML}{E6F5F1}
\definecolor{ybtintb}{HTML}{F3FBF9}
\definecolor{ybneg}{HTML}{B4442F}
\definecolor{ybgrey}{HTML}{8A8F98}

\makeatother

\usepackage{graphicx}

\usepackage{amsmath, amssymb}
\usepackage{booktabs}
\usepackage{multirow}

\usepackage{algorithm}
\usepackage{algpseudocode}
\usepackage{pifont} 

\newcommand{\cmark}{\ding{51}} 
\newcommand{\xmark}{\ding{55}} 
\usepackage{bbm}
\usepackage{tikz}
\newlength{\barwidth}
\usepackage{amssymb}
\usepackage{pifont}

\usepackage{lipsum}

\usepackage{hyperref}
\usepackage{iftex}
\ifXeTeX
    \usepackage{xeCJK}
\else
    \usepackage{CJKutf8}
    
\fi

\title{
Omni-Life: Benchmarking Omnimodal Real-Time AI Assistants in People's Daily Lives
}

\author{%
\makebox[\linewidth][c]{%
    \textbf{Mengze Hong}$^{1,2,*}$,
    \textbf{Ziwei Shang}$^{1,*}$,
    \textbf{Di Liang}$^{2,\dagger}$,
    \textbf{Zijian Li}$^{2}$
}\\[1pt]
\makebox[\linewidth][c]{%
    \textbf{Liang Zhu}$^{2}$,
    \textbf{Hongyu Shan}$^{2}$,
    \textbf{Robin Zhou}$^{2}$,
    \textbf{Minlong Peng}$^{3}$
}\\[1pt]
\makebox[\linewidth][c]{%
    \textbf{Zhiqiang Kou}$^{1}$,
    \textbf{Jia Wang}$^{4}$,
    \textbf{Di Jiang}$^{1}$,
    \textbf{Chen Jason Zhang}$^{1,\dagger}$
}\\[7pt]
\makebox[\linewidth][c]{%
    \small
    $^{1}$Hong Kong Polytechnic University
    \hspace{10pt}%
    $^{2}$Today AI
    \hspace{10pt}%
    $^{3}$Fudan University
}\\[1pt]
\makebox[\linewidth][c]{%
    \small
    $^{4}$Beijing University of Posts and Telecommunications
}\\[5pt]
\makebox[\linewidth][c]{%
    \footnotesize
    $^{*}$Equal contribution.
    \hspace{12pt}%
    $^{\dagger}$Corresponding authors.
}\\[5pt]
\makebox[\linewidth][c]{%
    \vspace{-2mm}
    \normalfont\small\sffamily
    \raisebox{-3pt}{%
        \includegraphics[height=11pt]{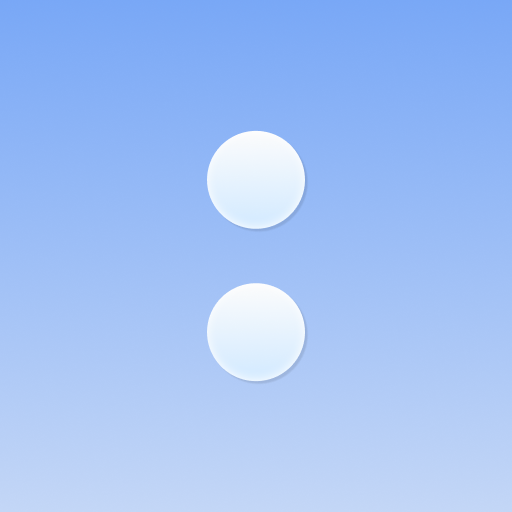}%
    }%
    \hspace{5pt}%
    \textcolor{repolink}{Project page}:\enspace
    \href{https://github.com/todayai-labs/omni-life}{%
        \textcolor{repolink}{https://github.com/todayai-labs/omni-life}%
    }%
}%
}

\begin{document}

\maketitle
\begin{abstract}
Real-time omnimodal assistants are expected to support people in everyday life, but audiovisual understanding alone does not establish whether they can provide timely and appropriate help. We introduce \textbf{Omni-Life}, a benchmark grounded in real-world user needs, deployment scenarios, and device-dependent observations. It combines controlled video synthesis for hazardous situations and edge cases with annotated real-world videos under a shared temporal protocol. The benchmark contains 400 videos: 290 requiring assistance, with 555 required intervention events, and 110 requiring silence throughout to test whether models can withhold unnecessary responses when no assistance is needed. We evaluate twelve configurations featuring mainstream real-time omnimodal models, jointly assessing event coverage, response content, timing, and unwanted output. At most 4.48\% of videos requiring assistance are handled completely correctly. To examine practical deployment strategies, we study task reminders and response-latency tolerance. Reminders can improve recall by 17.12 percentage points but add approximately 4.6 unwanted responses per minute of reference silence. Increasing latency tolerance by 10s improves recall by at most 5.95 percentage points. These results show that existing models remain insufficient to deliver reliable assistance in real-life applications, calling for tailored model optimization in instruction following, timely proactive intervention, and withholding responses when assistance is not needed.
\end{abstract}

\section{Introduction}

\begin{wrapfigure}{r}{7.47cm}
    \centering
    \includegraphics[width=0.85\linewidth]{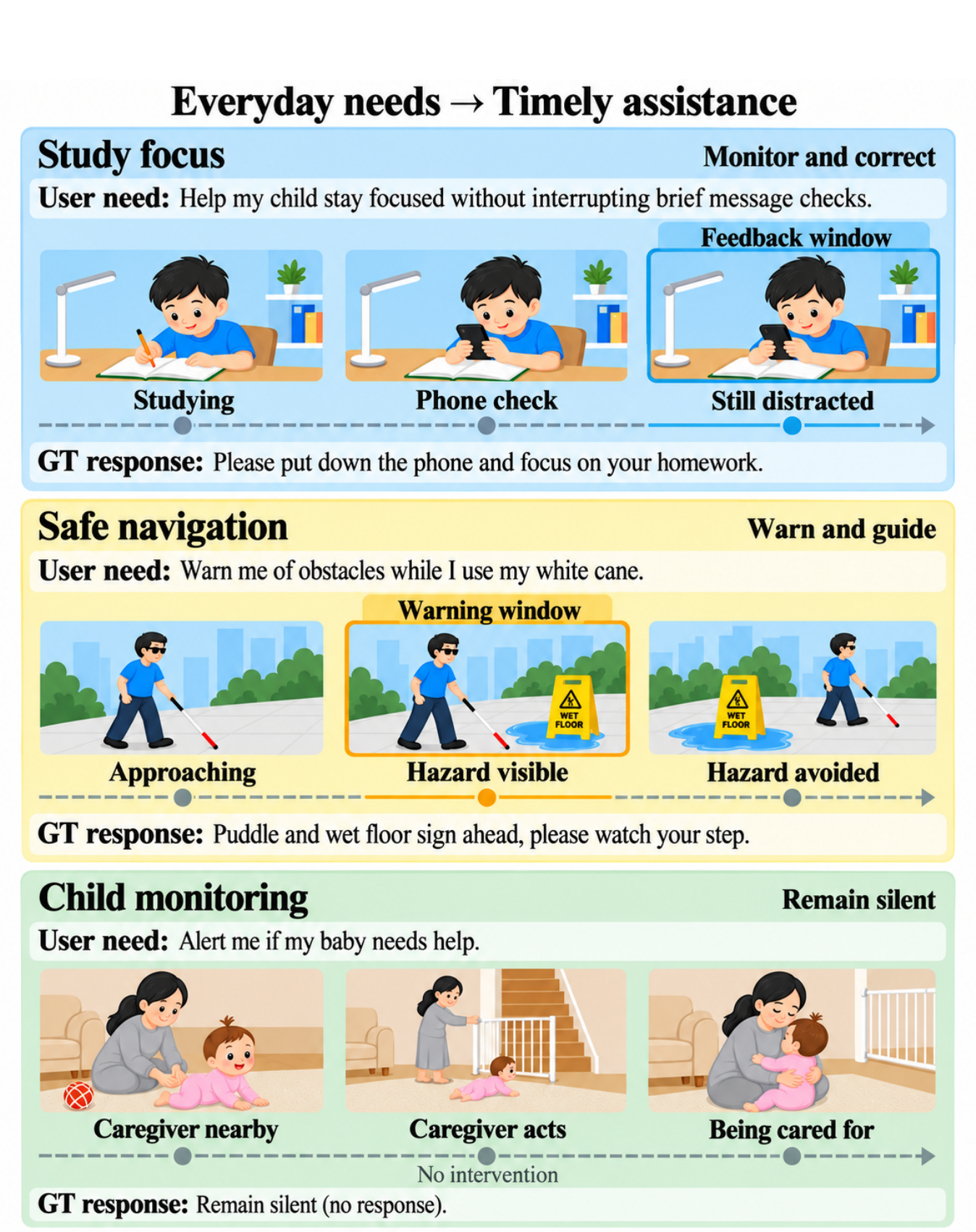}
    \caption{Illustration of real-life use cases for omnimodal real-time AI assistants.}
    \label{fig:need-grounded-motivation}
\end{wrapfigure}

Recent advances in omnimodal full-duplex foundation models are shifting human-AI interaction from turn-based conversations toward continuous, real-time assistance with audio-visual perceptions. This transition fundamentally changes what an assistant is expected to do. Rather than simply generating an answer after receiving a complete input, a real-time assistant must decide whether to respond, when to respond, and what response is appropriate as the surrounding environment changes \citep{lu2026omniinteractbenchmarkingrealworldstreaming}. While current efforts still largely emphasize model development and individual interaction capabilities, emerging demonstrations from mainstream AI products, such as SeedRealtime in Doubao and GPT-Live, already point toward a much broader future in which such models operate across different devices, environments, and user populations, ideally participating in people's everyday activities. This expectation also introduces new forms of failure: a model may interrupt when no help is needed, overlook an important event, or even provide incorrect information that either leads to minor user dissatisfaction or consequential safety risks. These transitions in how AI serves people create an emerging need for model evaluation beyond atomic abilities: evaluating in real-world applications is both timely and necessary to guide models toward truly helpful assistants in complex daily scenarios. Figure~\ref{fig:need-grounded-motivation} illustrates three everyday assistance needs and their appropriate responses.

A broad categorization of benchmark resources can be divided into capability-driven and downstream task-driven. On the one hand, existing omnimodal benchmarks evaluate streaming perception, audiovisual understanding, interaction triggers, response timing, and other capabilities required for real-time interaction, but typically do so within predefined observatory tasks or controlled interaction settings~\citep{lin2026streamingbench,Niu_2025_CVPR,Wang_2025_CVPR}. On the other hand, foundation model benchmarks have long expanded evaluation across domains, tasks, and skills, but their notion of scenario diversity is largely designed around conventional input-output systems and does not reflect the complexity introduced by real-time omnimodal interaction. In practice, appropriate model behavior depends on a compound of \emph{where} the interaction takes place, \emph{who} the model is serving, \emph{what} the deployed device can observe, and \emph{when} intervention is actually useful. A meaningful test may therefore be less about solving an exceptionally difficult maths problem, and more about whether a model can appropriately help a primary school student with homework, assist a blind person in navigating safely, or provide timely support during an unfolding daily activity. We therefore argue that evaluating omnimodal real-time interaction requires bridging these two perspectives together: the interaction capabilities and the broad, practical service scenarios in which these models play meaningful roles in everyday life, asking not only what models are capable of doing, but also what services they can reliably provide, for whom, under what conditions, and when.

In this paper, we introduce \textbf{Omni-Life}, to our knowledge the first real-time omnimodal benchmark centered on everyday human support needs. We organize it around a three-dimensional grid jointly conditioned on \textbf{scenario $\times$ population $\times$ device}: a scenario describes what is happening, a population identifies whose needs matter, and a device determines what the assistant can observe and whether the viewpoint is static or evolving. Rather than combining these categories mechanically, we use valid combinations to define concrete service tasks, in which people's actual needs and demands are valued over synthesized test cases solely for observational capabilities. This is further emphasized by paired test instances given the treatment of the user population, where the assistant should react meaningfully at the right moment or remain silent depending on the population that appears in the event.

The main contributions of this paper include:

\begin{enumerate}
    \item We develop an evaluation framework that connects real-world user needs with deployment scenarios and the observations available to different devices. The framework translates these factors into concrete service tasks with explicit conditions for intervention and appropriate silence, assessing model capabilities in real-life scenarios.
    
    \item We construct the \textbf{Omni-Life} benchmark, featuring 400 human-verified cases that combine controlled video synthesis of hazardous situations and edge cases with annotated real-world videos. A unified temporal annotation protocol specifies required assistance, response windows, and intervals of expected silence as the desired model behaviors.
    
    \item We perform a comprehensive evaluation of twelve real-time omnimodal model configurations through continuous streaming and turn-based QA, complemented by subgroup analyses, failure taxonomy, and controlled ablations. Our results reveal systematic limitations in timely and selective assistance, uncover trade-offs between intervention coverage and unwanted responses, and identify key challenges in reliable real-world deployment.
\end{enumerate}

\begin{table}[!t]
\centering
\caption{Comparison of benchmark tasks and evaluation protocols.}
\label{tab:benchmark-comparison}

\begingroup
\fontsize{9}{11}\selectfont
\setlength{\tabcolsep}{1pt}
\renewcommand{\arraystretch}{1.15}
\renewcommand{\tabularxcolumn}[1]{m{#1}}

\begin{tabularx}{\linewidth}{
>{\raggedright\arraybackslash}m{0.21\linewidth}
>{\raggedright\arraybackslash}m{0.15\linewidth}
>{\raggedright\arraybackslash}X
*{5}{c}
}
\toprule
\textbf{Benchmark}
& \textbf{Task focus}
& \textbf{Video source}
& \makecell{\textbf{Proactive}\\\textbf{response}}
& \makecell{\textbf{Response}\\\textbf{timing}}
& \makecell{\textbf{User}\\\textbf{groups}}
& \makecell{\textbf{Hazard}\\\textbf{alerts}}
& \makecell{\textbf{Silence-only}\\\textbf{videos}} \\
\midrule

\makecell[l]{StreamingBench\\
\citep{lin2026streamingbench}}
& \makecell[l]{Streaming\\video QA}
& YouTube
& \cmark & \cmark & \xmark & \xmark & \xmark \\
\addlinespace[4pt]

\makecell[l]{OVO-Bench\\
\citep{Niu_2025_CVPR}}
& \makecell[l]{Online\\video QA}
& \makecell[l]{Curated +\\YouTube}
& \cmark & \cmark & \xmark & \xmark & \xmark \\
\addlinespace[4pt]

\makecell[l]{OmniMMI\\
\citep{Wang_2025_CVPR}}
& \makecell[l]{Multimodal\\interaction}
& \makecell[l]{Curated +\\YouTube}
& \cmark & \cmark & \xmark & \cmark & \xmark \\
\addlinespace[4pt]

\makecell[l]{ProactiveVideoQA\\
\citep{wang2025proactivevideoqacomprehensivebenchmarkevaluating}}
& \makecell[l]{Proactive\\video QA}
& \makecell[l]{Public\\datasets}
& \cmark & \cmark & \xmark & \cmark & \xmark \\
\addlinespace[4pt]

\makecell[l]{PhoStream\\
\citep{lu2026phostreambenchmarkingrealworldstreaming}}
& \makecell[l]{Mobile\\assistance}
& \makecell[l]{Web + curated\\+ recordings}
& \cmark & \cmark & \xmark & \xmark & \xmark \\
\addlinespace[4pt]

\makecell[l]{OmniInteract\\
\citep{lu2026omniinteractbenchmarkingrealworldstreaming}}
& \makecell[l]{Spoken\\interaction}
& \makecell[l]{Recordings +\\curated}
& \cmark & \cmark & \xmark & \xmark & \xmark \\
\addlinespace[4pt]

\makecell[l]{OmniPro\\
\citep{zhao2026omnipro}}
& \makecell[l]{Proactive\\video tasks}
& \makecell[l]{LongVALE\\+ COIN}
& \cmark & \cmark & \xmark & \xmark & \xmark \\
\addlinespace[4pt]

\makecell[l]{EgoSocial\\
\citep{wang2025egosocialbenchmarkingproactiveintervention}}
& \makecell[l]{Social\\intervention}
& Ego4D
& \xmark & \xmark & \xmark & \xmark & \xmark \\

\midrule
\rowcolor{ybtint}
\textbf{Omni-Life}
& \makecell[l]{\textbf{Everyday}\\\textbf{assistance}}
& \makecell[l]{\textbf{Real +}\\\textbf{Synthetic}}
& \cmark & \cmark & \cmark & \cmark & \cmark \\
\bottomrule
\end{tabularx}

\endgroup
\end{table}

\section{Related Work}

\subsection{Omnimodal and Full-Duplex Models}

Multimodal foundation models are increasingly evolving from offline perception toward unified, streaming interaction. Recent omnimodal models integrate vision, audio, speech, and text within a single system, enabling continuous perception and low-latency response generation~\citep{xu2025qwen25omnitechnicalreport,cui2026minicpmo45realtimefullduplex}. In parallel, full-duplex spoken dialogue models such as Moshi relax the conventional turn-taking assumption by allowing a model to listen and speak simultaneously, supporting interruptions, backchannels, and overlapping speech~\citep{défossez2024moshispeechtextfoundationmodel}. More recent systems further combine full-duplex communication with continuous visual perception and proactive behavior, allowing models to see, listen, and speak while an environment is evolving~\citep{cui2026minicpmo45realtimefullduplex}. Together, these advances are shifting multimodal AI from static input-output processing toward real-time interaction, where model behavior must be continuously grounded in both the user and the surrounding environment.

\subsection{Benchmarking Real-time Interaction}

A growing body of benchmarks has begun to evaluate capabilities required by this interaction paradigm. StreamingBench and OVO-Bench study online video understanding and timestamp-aware reasoning over streaming visual inputs~\citep{lin2026streamingbench,Niu_2025_CVPR}. OmniMMI extends this setting toward multimodal streaming understanding and proactive reasoning~\citep{Wang_2025_CVPR}, while ProactiveVideoQA explicitly evaluates both response content and response timing in proactive video interaction~\citep{wang2025proactivevideoqacomprehensivebenchmarkevaluating}. Full-Duplex-Bench focuses on pause handling, backchanneling, turn-taking, and interruption management in spoken dialogue~\citep{lin2025full-duplex-bench}, and more recent benchmarks such as Omni-DuplexEval and OmniInteract evaluate when and what models should respond under real-time multimodal streams~\citep{he2026omniduplexevalevaluatingrealtimeduplex,lu2026omniinteractbenchmarkingrealworldstreaming}. PhoStream further evaluates mobile assistants through on-screen and off-screen streaming QA, including egocentric videos from blind users~\citep{lu2026phostreambenchmarkingrealworldstreaming}. Building on existing protocols for online perception and interaction, Omni-Life organizes service tasks across population, scenario, and device with explicit intervention and silence requirements grounded in user needs and observable evidence, as compared with representative benchmarks in Table~\ref{tab:benchmark-comparison}.

\subsection{Application-Oriented Benchmarks}
Another line of work moves evaluation from generic capabilities toward realistic application contexts. WorkArena evaluates agents on routine knowledge-work tasks in enterprise software~\citep{pmlr-v235-drouin24a}, while OSWorld studies open-ended computer-use tasks across real desktop applications~\citep{NEURIPS2024_5d413e48}. Domain-oriented benchmarks such as QualBench further evaluate whether general-purpose LLMs possess the localized professional knowledge required across multiple real-world vertical domains~\citep{hong-etal-2025-qualbench}. EgoSocial evaluates socially appropriate intervention decisions from egocentric audiovisual cues in short social-interaction clips~\citep{wang2025egosocialbenchmarkingproactiveintervention}. OmniSmartHome evaluates grounding and executing spoken requests using visual and spatial-audio context in synthetic and real smart-home episodes~\citep{jung2026omnismarthomemultimodalreasoningbenchmark}. These efforts demonstrate the value of evaluating models according to the environments and services in which they are expected to operate. Omni-Life evaluates a complementary setting of \emph{everyday-life interaction} with real-time omnimodal assistants, where the model may need to continuously observe an unfolding situation, determine whether assistance is needed, identify whom it should assist, and respond within an appropriate temporal window. We therefore organize evaluation jointly across scenario, population, device, and time, treating the real-world service provided by an omnimodal assistant as the basic unit of evaluation.

\section{Omni-Life Dataset}
\label{sec:omnilife}

\subsection{Overview}
\label{sec:dataset_overview}

The goal of Omni-Life is to study how future AI systems can support people in everyday life across heterogeneous user groups, real-world scenarios, and deployment devices. Unlike %
\begin{wrapfigure}{r}{7.47cm}
    \centering
    \includegraphics[width=\linewidth]{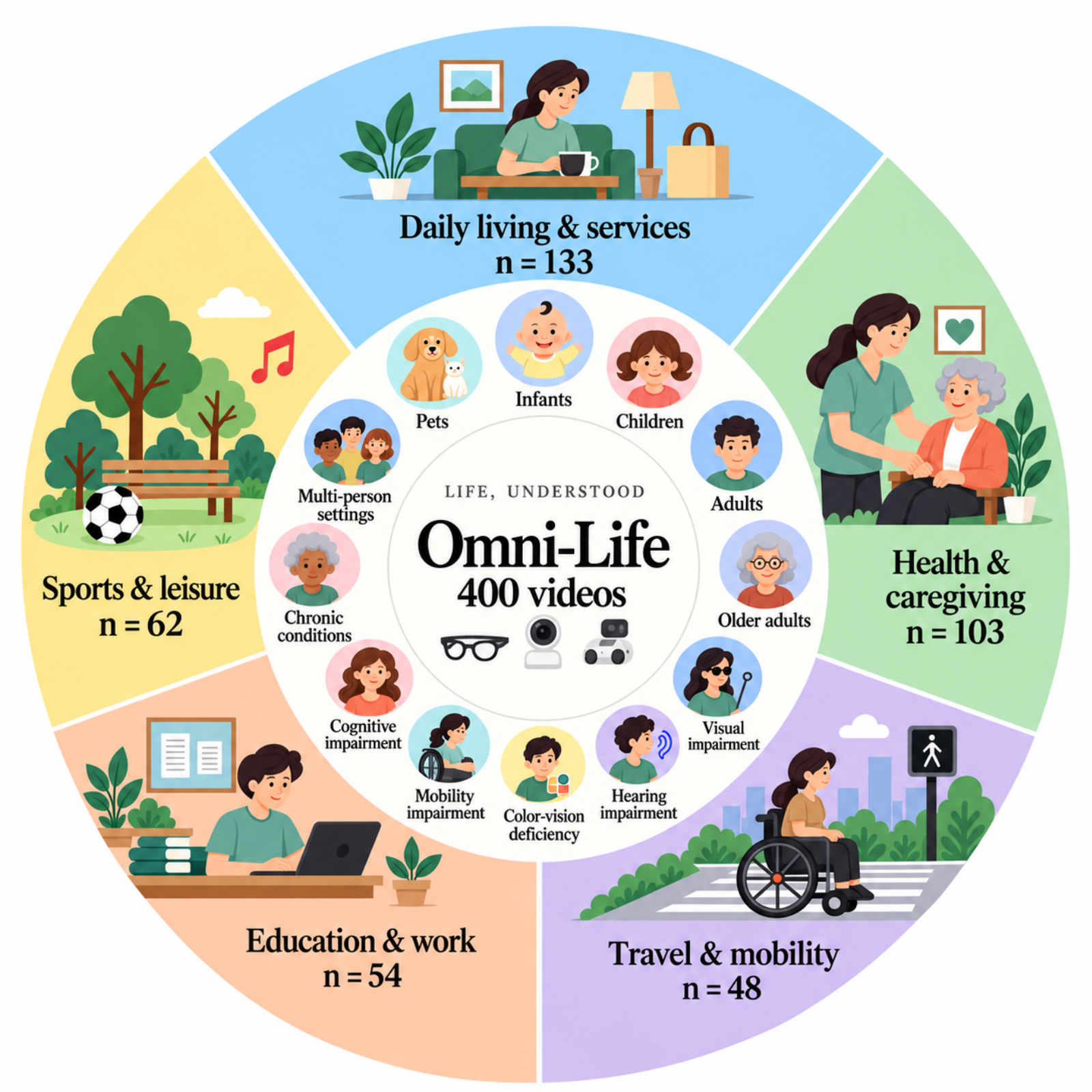}
    \caption{Five scenario domains and twelve service groups. Counts are videos; rings encode separate dimensions with schematic angles.}
    \label{fig:benchmark-coverage}
\end{wrapfigure}
conventional multimodal benchmarks that primarily evaluate whether a model can recognize, describe, or answer questions about observed content, Omni-Life focuses on a more fundamental problem: \emph{when should an AI respond to an ongoing real-world situation, and what behavior is appropriate in that situation?} Figure~\ref{fig:benchmark-coverage} summarizes the scenario and service-group coverage, with device icons illustrating observation contexts. We formulate this decision by jointly considering three aspects:
\begin{equation}
    f:(P,S,D)\rightarrow A,
\end{equation}
where $P$ denotes user or population characteristics, $S$ denotes the current scenario, $D$ denotes the available sensing and interaction devices, and $A$ denotes the expected AI behavior. We define five behavior labels: \texttt{[warn]} for alerting users to potential risks, \texttt{[inform]} for proactively providing relevant information, \texttt{[feedback]} for evaluating or correcting user actions, \texttt{[answer]} for responding to direct user queries, and \texttt{[wait]} for situations where no response is appropriate. Together, these behaviors capture both when AI assistance is needed and when the appropriate behavior is to remain silent, reflecting the dynamic nature of real-life interactions and the importance of timely, context-aware assistance.

\begin{figure}[!t]
    \centering
    \includegraphics[width=\linewidth]{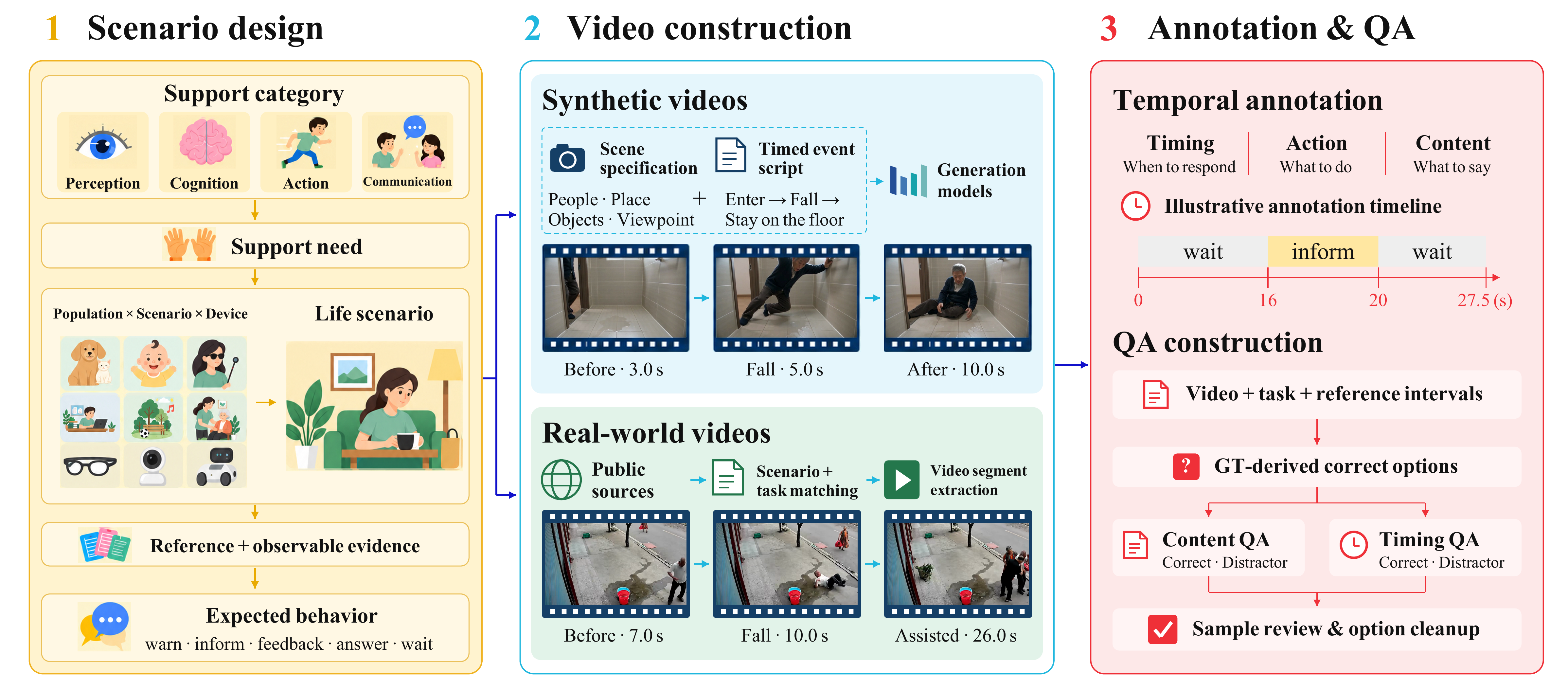}
    \caption{Construction pipeline of Omni-Life. Support needs guide scenario design and video sourcing through controlled synthesis or selection from public real-world sources. Temporal annotations define response windows, behavior labels, and reference content, and provide correct options for turn-based QA.}
    \label{fig:dataset_pipeline}
\end{figure}

Figure~\ref{fig:dataset_pipeline} summarizes the construction pipeline of Omni-Life. We first translate support needs into life scenarios and reference-grounded assistant behaviors. Videos are either synthesized from structured specifications or selected from public real-world sources. Quality checks cover observable evidence, temporal consistency,
and answer leakage. Videos include task queries and timed action, response,
and silence annotations for streaming evaluation and turn-based QA.
\subsection{Systematic Scenario Construction}
\label{sec:scenario_construction}

A central challenge in constructing Omni-Life is deciding which situations in everyday life genuinely call for AI support. Rather than directly enumerating activities, we organize the scenario space hierarchically as
\begin{equation}
    \text{Support Category}
    \rightarrow
    \text{Support Need}
    \rightarrow
    \text{Life Scenario}.
\end{equation}
This organization is motivated by Gibson's notion of \emph{affordance}, which emphasizes that the opportunities and constraints presented by an environment depend on the capabilities of the person interacting with it \citep{Gibson_2014}. The same staircase, for example, may be an ordinary path for a walking adult but an obstacle for a wheelchair user. From the perspective of AI assistance, the relevant question is therefore not simply ``what is happening in the scene?'', but ``\textbf{what support does this person need in this situation?}''

To avoid defining such needs solely through intuition, we further draw on the WHO International Classification of Functioning, Disability and Health (ICF), which relates human functioning to activities, participation, and environmental factors\footnote{WHO ICF: \href{https://www.who.int/classifications/international-classification-of-functioning-disability-and-health}{International Classification of Functioning, Disability and Health}.}. Following this perspective, we organize everyday support into four broad categories:
\begin{enumerate}
    \item \textbf{Perception}: access to visual, auditory, color, and other environmental information that may be difficult to obtain directly;
    \item \textbf{Cognition}: awareness of state changes, situational understanding, judgment, attention, and memory;
    \item \textbf{Action}: mobility, action execution, and assistance when independent response is difficult;
    \item \textbf{Communication}: expression and understanding of participants, targets, and states in multi-party interactions.
\end{enumerate}
Importantly, user characteristics provide context for these support needs but do not directly determine the expected AI behavior. The same user may require different behaviors depending on the situation: for example, a visually impaired user may require \texttt{[inform]} when identifying an object, \texttt{[answer]} after explicitly asking about the environment, \texttt{[warn]} when approaching an immediate hazard, and \texttt{[wait]} when no assistance is needed.

Population, scenario, and observation conditions are specified jointly when constructing a case. A proposed response must be justified by evidence available from the selected viewpoint: a close-up may identify an object but cannot establish what is happening outside its field of view. The construction process filters incompatible combinations of population, behavior, viewpoint, and task form before expanding a candidate into a scene specification. The device dimension guides camera placement and the evidence that the scene must make observable.

\subsubsection{Reference-Grounded Behavior Design}
\label{sec:behavior_design}

The life scenarios above specify what kinds of situations Omni-Life should cover, but they do not yet determine \emph{how the AI should behave} in each situation. To make this decision, we further identify the reference against which the observed situation should be interpreted. For skill-related scenarios, AI can support users in two different ways: reporting \emph{what happened} and providing feedback on \emph{how an action was performed}. This distinction draws on \emph{knowledge of results} (KR) and \emph{knowledge of performance} (KP) in motor learning \citep{Salmoni_1984}. We therefore use \texttt{[inform]} to report states or outcomes, and \texttt{[feedback]} to evaluate or correct the action itself. When an incorrect action further creates an immediate risk of harm, the appropriate behavior escalates from \texttt{[feedback]} to \texttt{[warn]}. Rule-governed outcomes are determined relative to constitutive rules \citep{Searle_1969}. In Omni-Life, \texttt{[inform]} or \texttt{[answer]} is appropriate only when the observed evidence supports the conclusion. Conversational cooperation emphasizes relevant contributions, while turn-taking concerns the allocation of speaking turns \citep{Grice_1975,Sacks_Schegloff_Jefferson_1974}. In Omni-Life, the AI should \texttt{[wait]} when it is not being called or when another speaker has not finished, making appropriate silence an explicit behavior rather than a missing response.

These principles turn each life scenario into an explicit intervention decision. For each scenario, we specify the reference required for judgment, the observable evidence supporting that judgment, and the corresponding AI behavior. We then construct cases around important decision boundaries: correct versus incorrect actions for \texttt{[feedback]}, ordinary versus hazardous states for \texttt{[inform]} or \texttt{[feedback]} versus \texttt{[warn]}, and addressed versus non-addressed interactions for \texttt{[answer]} versus \texttt{[wait]}. In this way, behavior labels are tied to observable conditions rather than assigned from the scenario category alone.

\subsection{Synthetic and Real-World Video Sourcing}
\label{sec:video_sourcing}

\paragraph{From life scenarios to observable events.}
After defining the appropriate behavior and its decision conditions, we translate each life scenario into concrete events that can be observed in videos. We represent the temporal progression as
\begin{equation}
    E=(e_1,e_2,\ldots,e_n),
\end{equation}
where each $e_i$ describes an observable event or state change relevant to the intervention decision. For example, a fall-risk scenario specifies the source of risk, how the person approaches it, and the event that makes the risk identifiable. This specifies the observable evidence required to justify the expected behavior. The same life scenario can be instantiated with different people, environments, objects, viewpoints, and event progressions, increasing visual and contextual diversity while preserving the underlying decision being evaluated. When multiple relevant events occur, we prioritize them according to urgency, with safety-critical events taking precedence over routine informational needs.

\paragraph{Synthetic video generation.}
We synthesize videos with Seedance video models\footnote{Official model catalogue: \url{https://seed.bytedance.com/en/models}.} from structured specifications that separate scene settings, timed events, and candidate response annotations. The generation prompt contains the location, participant appearances, key objects, camera viewpoint, and event descriptions; reference assistant replies remain in the annotation record. For cases triggered by a spoken question, prompts retain that utterance and environmental sound while prohibiting an answering voice, explanatory narration, and subtitles. This preserves the distinction between the evidence for a decision and the response being evaluated.

Event descriptions specify an ordered progression with visible state changes and pauses between repeated actions. Counting cases require distinguishable completed actions, while longer sequences can defer a question until after intervening events. Paired specifications share the setting and most events while changing a condition relevant to intervention, such as whether a hazard remains unhandled or is resolved by another person. They provide intervention and non-intervention cases in similar everyday situations without treating a scene category alone as permission to respond.

The generation guidelines favor visually explict actions and state changes, such as falling or turning off a stove, over subtle deviations of otherwise normal actions. Missing or ambiguous decision-critical evidence is a criterion for regenerating, reformulating, or excluding a candidate. These criteria aim to reduce the risk of conflating synthesis defects with model errors.

\paragraph{Real-world video sourcing.}
To complement controlled synthesis with naturally occurring variation, we additionally source publicly accessible real-world videos that instantiate the predefined life scenarios. These videos introduce natural diversity in human behavior, environments, camera motion, and event progression that may not be fully captured by generated content. Synthetic and real-world videos are subsequently annotated under the same behavior and temporal framework, providing a common evaluation space across both sources. Each selected clip is treated as a separate evaluation video, with temporal labels expressed on the resulting clip's timeline.

\paragraph{Quality control.}
A specification linter supports checks before rendering by flagging overly short event segments, descriptions with no observable action, and descriptions that restart the setting instead of continuing the preceding event. Reusable prompt blocks specify stable participant appearances, consistent object counts, camera continuity, and physically coherent motion and contact. Prompt constraints guide generation; the rendered clip still requires review. Video quality control focuses on the visibility of decision-critical evidence, temporal consistency, and absence of answer leakage; synthetic clips additionally require physical plausibility and participant and object consistency. Additional construction and quality-control criteria are provided in Appendix~\ref{app:data_construction}.

\subsection{Temporal Annotation and QA Construction}
\label{sec:temporal_annotation}

Each video is paired with a task query and an ordered sequence of reference intervals. An interval records its start and end times, one of the five behavior labels, and Chinese and English reference content where a response is required. The two languages express the same assistance requirement. The half-open interval $[s,e)$ defines when that assistance is appropriate; it can differ from the period during which an object is merely visible. Explicit \texttt{[wait]} intervals have empty response content and cover periods when no intervention is appropriate.
\begin{wraptable}{r}{7.47cm}
\centering
\caption{Omni-Life dataset statistics.}
\label{tab:dataset_statistics}
\small
\setlength{\tabcolsep}{3pt}
\renewcommand{\arraystretch}{1.15}
\arrayrulecolor{black}

\begin{tabularx}{\linewidth}{
  @{}>{\raggedright\arraybackslash}X r r r@{}
}
\toprule
\textbf{Statistic}
  & \textbf{Synthetic}
  & \textbf{Real-world}
  & \textbf{Total} \\
\midrule

\textbf{Videos}
  & \textbf{277} & \textbf{123} & \textbf{400} \\
\quad Intervention
  & 184 & 106 & 290 \\
\quad Remain Silent
  & 93 & 17 & 110 \\
\midrule

\textbf{Intervention nodes}
  & \textbf{255} & \textbf{300} & \textbf{555} \\
\quad Inform
  & 139 & 223 & 362 \\
\quad Warn
  & 92 & 57 & 149 \\
\quad Feedback
  & 11 & 18 & 29 \\
\quad Answer
  & 13 & 2 & 15 \\
\midrule

\textbf{Silence intervals}
  & \textbf{370} & \textbf{315} & \textbf{685} \\
\midrule

\textbf{QA questions}
  & \textbf{554} & \textbf{246} & \textbf{800} \\
\bottomrule
\end{tabularx}
\end{wraptable}
We then create turn-based QA and each video has one content question and one timing question. Content questions ask what the assistant should say at a given moment or over the full video. Timing questions ask when it should respond.
Question stems use predefined templates, and correct options come from the annotations. Gemini generates incorrect content options related to the same scenario and task. Incorrect timing options are generated by rules using alternative time windows or start times. For videos requiring silence throughout,
the correct answer is to remain silent.

Finally, we manually review a sample of the questions, balance their wording and length, and remove duplicate options. The final set contains 800 questions: 400 content questions and 400 timing questions.
Further details are provided in Appendix~\ref{app:data_construction}.

\subsection{Dataset Composition}
\label{sec:dataset_composition}
The evaluation set contains \textbf{400 videos}: \textbf{277 synthesized} and \textbf{123 real-world} clips. Of these, 290 require at least one intervention and 110 require silence throughout their annotated timelines. The annotations contain \textbf{555 required intervention nodes} and 685 \texttt{[wait]} intervals, including quiet periods within intervention videos. Video counts and interval counts therefore describe different units. 
Table~\ref{tab:dataset_statistics} separates the composition by video source. The required interventions comprise 362 \texttt{[inform]}, 149 \texttt{[warn]}, 29 \texttt{[feedback]}, and 15 \texttt{[answer]} instances. The five behavior labels are not forced into equal proportions because their applicability depends on the task. The evaluated videos cover 12 user and service groups, with five application domains (see Figure \ref{fig:benchmark-coverage}). Records retain population, scene, viewpoint, and task-derived assistant-role metadata for descriptive analysis.

\section{Results and Discussion}
\label{sec:results}

\subsection{Experimental Setup}
\label{sec:experimental_setup}

\paragraph{Models and deployment.}
We evaluate eight locally deployed open-weight systems and four API configurations in both streaming and turn-based QA (Table~\ref{tab:model_setup}). Local systems use released checkpoints and inference implementations on an eight-H200 server, with isolated sessions and native memory and context management. MOSS Preview uses the real-time SFT checkpoint; \mbox{DuplexOmni} uses only the S1 Thinker, without its S2 reasoning layer. API models use Google Live\footnote{Gemini 3.8 Audio model card: \url{https://deepmind.google/models/model-cards/gemini-3-8-audio/}.} or Alibaba Omni Realtime endpoints.

\begin{table}[!t]
\definecolor{modelLocalBg}{HTML}{E2E6ED}

\centering
\caption{Evaluated local and API-based model interfaces.}
\label{tab:model_setup}
\small
\setlength{\tabcolsep}{3pt}
\renewcommand{\arraystretch}{1.12}
\begin{tabular}{@{}>{\raggedright\arraybackslash}p{0.49\linewidth}>{\raggedright\arraybackslash}p{0.48\linewidth}@{}}
\toprule
\textbf{Model} & \textbf{Serving / audio input} \\
\midrule

\multicolumn{2}{c}{
  \cellcolor{modelLocalBg}
  \textbf{Open-weight and Locally deployed Models}
} \\
\multicolumn{2}{@{}l@{}}{\textbf{Open-weight models, locally deployed}} \\
MiniCPM-o 4.5~\citep{cui2026minicpmo45realtimefullduplex} & Native duplex / audio encoder \\
MOSS-VL~\citep{wang2026mossvltechnicalreport} & Streaming service / SenseVoice \\
MOSS Preview~\citep{wang2026mossvideopreviewrealtimevideounderstanding} & Streaming service / SenseVoice \\
Venus-Omni~\citep{antgroup2026realtimevenusfullduplexinteractionasynchronous} & Native duplex / audio encoder \\
Gander~\citep{orantqing2026multimodalduplexinteractionagent} & OnlineRunner / audio encoder \\
AURA~\citep{lu2026auraalwaysonunderstandingrealtime} & vLLM / Qwen3-ASR-1.7B \\
JoyAI~\citep{yao2026joyaivlinteractionrealtimevisionlanguageinteraction} & StreamingInferAdapter / Qwen3-ASR-1.7B \\
DuplexOmni (S1)~\citep{huang2026duplexomnirealtimelisteningseeing} & vLLM Thinker / audio encoder \\
\midrule

\multicolumn{2}{c}{
  \cellcolor{modelLocalBg}
  \textbf{API-based Models}
} \\
\multicolumn{2}{@{}l@{}}{\textbf{API-based models}} \\
Gemini 3.8 Live & Google Live / native audio \\
Qwen3.5-omni-flash-realtime~\citep{qwenteam2026qwen35omnitechnicalreport} & Alibaba Realtime / native audio \\
Qwen3.5-omni-plus-realtime~\citep{qwenteam2026qwen35omnitechnicalreport} & Alibaba Realtime / native audio \\
Qwen3.8-omni-flash-realtime~\citep{qwenteam2026qwen38omninativeomnimodalagents} & Alibaba Realtime / native audio \\
\bottomrule
\end{tabular}
\end{table}

\paragraph{Streaming protocol.}
Each session receives the Chinese benchmark instruction, a written task, and source video, without reference answers or annotation windows. Media follow the source time without future inputs: frames are sampled at a nominal 1~FPS with source timestamps, and audio is 16-kHz mono PCM. The 30 videos without source audio receive no artificial track. All main-experiment models, including Gemini and Qwen, receive text task prompts. For both MOSS models, AURA, and JoyAI, speech in the source
video is transcribed with ASR and supplied as text, because
these configurations do not process audio directly.
Both MOSS models use SenseVoice~\citep{an2024funaudiollmvoiceunderstandinggeneration},
while AURA and JoyAI use Qwen3-ASR-1.7B~\citep{shi2026qwen3asrtechnicalreport}.The instruction requests \texttt{[tag]} response text when intervention is needed and no output for \texttt{[wait]}. Local systems and Qwen return text; Gemini returns audio with transcription. Interface-specific scheduling and response grouping are retained; decoding, media delivery, and recording details are in Appendix~\ref{app:inference_configuration}.

\paragraph{Turn-based QA.}
Each video supplies one content question and one timing question. Media are truncated at the question's observation cutoff when specified; otherwise, the full clip is available. Models are instructed to return option letters only, separated by semicolons for multiple choice. QA decoding settings are given in Appendix~\ref{app:inference_configuration}.

\paragraph{Scoring and metrics.}
DeepSeek-V4-Flash-0731~\citep{deepseekai2026deepseekv4highlyefficientmilliontoken} evaluates each answer against the task- and node-specific reference using a binary true/false judgment. Semantically equivalent and minimally sufficient answers are accepted, while answers that omit required information or contain substantive contradictions are rejected. The model does not access the video or evaluate response timing, and correctness of behavior tags is not required. A prediction is considered a match only if its content score is 1 and its onset falls within the original $[s,e)$ window. We use maximum-cardinality, order-preserving, one-to-one matching, without additional timing tolerance. These judgments are conditioned on the annotated ground-truth references and do not independently verify facts against the video (Appendix~\ref{app:content_matching}).

Let $G=555$ be the required-node count, $P$ the response-event count across all 400 videos, and $TP$ the timely, content-correct matches. \textbf{Intervention Recall} is $TP/G$ and \textbf{Intervention F1} is $2TP/(P+G)$; unmatched outputs and nodes are false positives and false negatives, respectively. \textbf{Full Correct} is the fraction of the 290 intervention videos with all nodes matched, no extra response or response-interval overlap with silence, and successful completion. \textbf{Silence Success} is the fraction of the 110 fully silent videos completed with no actual output. QA \textbf{Content Acc.} and \textbf{Timing Acc.} use exact answer-set matching over 400 questions each; \textbf{Joint Acc.} requires both answers for a video to be correct. Missing, invalid, and failed QA answers receive no credit, and failed streaming sessions remain in the denominators. F1 uses interface-specific response units and should be read with the video-level metrics; Appendix~\ref{app:silence_metrics} gives further scoring details.

\paragraph{Human baseline.}
Four human annotators independently completed both the turn-based QA and streaming evaluations. Reported scores are the unweighted means of the four participants' accuracies. Streaming responses were scored by a judge for semantic correctness and temporal alignment with reference interventions. Implementation details, exclusions, and score aggregation are provided in Appendix~\ref{app:human_baseline}.

\begin{table*}[!t]
\definecolor{mainGroup}{HTML}{E2E6ED}

\definecolor{mainHighlight}{HTML}{C1DFDA}
\definecolor{underlineGreenLight}{HTML}{E5F1EF}
\newcommand{\mainBest}[1]{%
  \cellcolor{mainHighlight}\textbf{#1}%
}
\newcommand{\mainSecond}[1]{%
  \cellcolor{underlineGreenLight}\underline{#1}%
}

\centering
\caption{Results on Omni-Life (\%; higher is better, $\uparrow$). Excluding the human baseline, the highest scores are marked in \protect\colorbox{mainHighlight}{\textbf{green}}
and the second-best scores are marked in \protect\colorbox{underlineGreenLight}{\underline{lighter green}}}
\label{tab:omnilife-main}

\fontsize{9}{11}\selectfont
\setlength{\tabcolsep}{3pt}
\renewcommand{\arraystretch}{1.25}
\arrayrulecolor{black}

\begin{tabular*}{\textwidth}{
  @{\extracolsep{\fill}}
  l r r r r
  @{\hspace{14pt}\extracolsep{\fill}}
  r r r
  @{}
}
\toprule

& \multicolumn{4}{c}{\textbf{Streaming}}
& \multicolumn{3}{c}{\textbf{Turn-based QA}} \\
\cmidrule(lr){2-5}
\cmidrule(lr){6-8}

& \multicolumn{1}{c}{
    \shortstack{\textbf{Full}\\\textbf{Correct} $\uparrow$}
  }
& \multicolumn{1}{c}{
    \shortstack{Intervention\\Recall $\uparrow$}
  }
& \multicolumn{1}{c}{
    \shortstack{Intervention\\F1 $\uparrow$}
  }
& \multicolumn{1}{c}{
    \shortstack{Silence\\Success $\uparrow$}
  }
& \multicolumn{1}{c}{
    \shortstack{\textbf{Joint}\\\textbf{Acc.} $\uparrow$}
  }
& \multicolumn{1}{c}{
    \shortstack{Content\\Acc. $\uparrow$}
  }
& \multicolumn{1}{c}{
    \shortstack{Timing\\Acc. $\uparrow$}
  } \\
\midrule

\multicolumn{8}{c}{
  \cellcolor{mainGroup}\strut
  \textbf{Human Baseline}
} \\

Human
& 53.10 & 62.34 & 64.49 & 91.82
& 60.38 & 82.13 & 66.13 \\
\midrule

\multicolumn{8}{c}{
  \cellcolor{mainGroup}\strut
  \textbf{Open-weight Models}
} \\

MOSS Preview
& 0.00 & 1.08 & 1.09 & 2.73
& 11.00 & 45.00 & 29.00 \\

AURA
& 0.00
& \mainSecond{12.61}
& 1.63 & 0.91
& 24.00
& \mainSecond{65.75}
& 31.00 \\

JoyAI
& 0.00
& \mainBest{20.00}
& 1.67 & 0.00
& 13.25 & 63.00 & 19.00 \\

DuplexOmni (S1)
& 0.69 & 4.50 & 4.67 & 0.00
& 2.75 & 12.25 & 24.25 \\

MiniCPM-o 4.5
& 1.38 & 1.44 & 2.63
& \mainSecond{95.45}
& 13.00 & 53.50 & 25.75 \\

Venus-Omni
& 2.07 & 1.98 & 3.18 & 77.27
& 13.25 & 58.25 & 26.00 \\

Gander
& 2.07 & 1.08 & 2.05
& \mainBest{99.09}
& 3.50 & 34.00 & 12.00 \\

MOSS-VL
& \mainBest{4.48}
& 5.77
& \mainBest{7.06}
& 82.73
& 5.75 & 18.50 & 35.00 \\
\midrule

\multicolumn{8}{c}{
  \cellcolor{mainGroup}\strut
  \textbf{API-based Models}
} \\

Qwen3.5 Plus
& 2.07 & 3.78 & 4.88 & 80.91
& 14.25 & 45.25 & 40.25 \\

Qwen3.8 Flash
& 2.76 & 2.70 & 3.98 & 84.55
& \mainSecond{32.00}
& 58.25
& \mainSecond{48.75} \\

Qwen3.5 Flash
& 3.10 & 4.14 & 5.39 & 80.91
& 17.25 & 48.50 & 30.25 \\

Gemini 3.8 Live
& \mainSecond{3.79}
& 3.42
& \mainSecond{5.45}
& 93.64
& \mainBest{42.25}
& \mainBest{78.25}
& \mainBest{51.25} \\
\bottomrule
\end{tabular*}
\end{table*}

\subsection{Main Results}
\label{sec:main_results}

Table~\ref{tab:omnilife-main} reports both streaming intervention and explicit turn-based question-answering performance. The results separate event coverage, useful content, and appropriate silence.

\paragraph{Reliable proactive assistance remains difficult.}
The strongest Full Correct score is 4.48\% among the twelve evaluated model configurations, followed by Gemini at 3.79\%. MOSS-VL also leads Intervention F1 at 7.06\%, but its recall is only 5.77\%. Low video-level correctness therefore accompanies low node coverage; it is not explained solely by requiring every event in a video to be handled correctly. Supporting streaming input does not by itself establish reliable execution of a standing service task.

\paragraph{Intervention and silence must be evaluated together.}
Table~\ref{tab:omnilife-main} shows two contrasting patterns. JoyAI achieves the highest Intervention Recall (20.00\%), but its F1 is only 1.67\% and its Full Correct is zero. Gander achieves 99.09\% Silence Success but only 1.08\% recall.

Table~\ref{tab:response-outcomes} provides further detail. For JoyAI, 74.6\% of required nodes are paired with an on-time but incorrect response; the corresponding rate for AURA is 83.4\%. For Gander, 93.5\% of required nodes have no later reply. Incorrect responses include missing information, unhelpful content, and factual errors, so they should not all be interpreted as hallucinations. Together, these results show why event coverage, response quality, and silence must be evaluated jointly.

\begingroup
\widowpenalty=10000
\paragraph{Explicit QA competence does not establish continuous assistance capability.}
Gemini achieves 78.25\% Content Accuracy in QA but only 3.42\% streaming Intervention Recall. Answer selection after a question differs from initiating and delivering assistance while a task remains active. 

\endgroup

\begin{table*}[!t]
\definecolor{outGroup}{HTML}{E2E6ED}

\centering
\caption{Response outcomes for all 555 required interventions, reported separately for each model (\%). The highest value in each \textbf{Correct} column is \textbf{bold}.}
\label{tab:response-outcomes}

\fontsize{9}{11}\selectfont
\setlength{\tabcolsep}{4pt}
\renewcommand{\arraystretch}{1.20}
\arrayrulecolor{black}

\begin{tabular*}{\textwidth}{
  @{\extracolsep{\fill}}
  l r r
  @{\hspace{12pt}\extracolsep{\fill}}
  r r
  @{\hspace{12pt}\extracolsep{\fill}}
  r r
  @{}
}
\toprule
\multicolumn{1}{c}{\multirow{2}{*}{\textbf{Model}}}
& \multicolumn{2}{c}{\textbf{On-time}}
& \multicolumn{2}{c}{\textbf{Late}}
& \multicolumn{1}{c}{\multirow{2}{*}{\shortstack{\textbf{No later}\\\textbf{reply}}}}
& \multicolumn{1}{c}{\multirow{2}{*}{\textbf{Others}}} \\
\cmidrule(lr){2-3}
\cmidrule(lr){4-5}
& \multicolumn{1}{c}{\textbf{Correct}$\uparrow$}
& \multicolumn{1}{c}{\textbf{Wrong}$\downarrow$}
& \multicolumn{1}{c}{\textbf{Correct}$\uparrow$}
& \multicolumn{1}{c}{\textbf{Wrong}$\downarrow$}
& & \\
\midrule
\multicolumn{7}{c}{\cellcolor{outGroup}\strut\textbf{Open-weight Models}} \\
MiniCPM-o 4.5
& 1.4
& 2.5
& 1.4
& 2.3
& 82.9
& 9.4 \\
MOSS-VL
& 5.8
& 10.1
& 2.5
& 8.5
& 66.1
& 7.0 \\
MOSS Preview
& 1.1
& 9.0
& 0.7
& 3.6
& 82.3
& 3.2 \\
Venus-Omni
& 2.0
& 7.7
& 1.4
& 3.1
& 80.5
& 5.2 \\
Gander
& 1.1
& 2.2
& 0.9
& 0.9
& 93.5
& 1.4 \\
AURA
& 12.6
& 83.4
& 0.2
& 2.5
& 0.9
& 0.4 \\
JoyAI
& \textbf{20.0}
& 74.6
& 0.4
& 1.6
& 3.2
& 0.2 \\
DuplexOmni (S1)
& 4.5
& 7.2
& 0.0
& 1.4
& 73.7
& 13.2 \\
\midrule
\multicolumn{7}{c}{\cellcolor{outGroup}\strut\textbf{API-based Models}} \\
Gemini 3.8 Live
& 3.4
& 4.0
& 2.7
& 3.4
& 83.8
& 2.7 \\
Qwen3.5 Plus
& 3.8
& 8.8
& \textbf{3.4}
& 3.6
& 67.4
& 13.0 \\
Qwen3.5 Flash
& 4.1
& 7.4
& 3.1
& 4.9
& 67.6
& 13.0 \\
Qwen3.8 Flash
& 2.7
& 7.0
& 3.1
& 1.6
& 71.4
& 14.2 \\
\bottomrule
\end{tabular*}
\end{table*}

\paragraph{Separating timing from content.}
Table~\ref{tab:response-outcomes} preserves the main-score matches, then pairs remaining nodes and responses one-to-one, prioritizing on-time incorrect responses, late correct responses, and late incorrect responses. Late means onset at or after the window end, without an upper bound on lateness. The four response categories distinguish on-time or late onset and correct or incorrect content. No later reply means no recorded response at or after the node onset in a completed session. Others retain failed sessions, responses already assigned elsewhere, and missing evidence. This diagnostic gives no additional main-score credit and does not establish which event a late reply was intended to address. Early outputs remain unmatched responses in the main F1 calculation.

\paragraph{Useful content can arrive after the opportunity to help.}
For example, in an elevator-monitoring video, MOSS-VL reports
that the door has opened at 24.18~s, after the required response
window of $[18.75,21.75)$~s. Section~\ref{sec:ablations}
examines whether extending response windows gives such
late responses credit under a relaxed timing criterion.

\subsection{Population, Scenarios, and Intervention Requirements}
\label{sec:subgroup_failure_analysis}

\paragraph{Population-level breakdown.}
Intervention Recall remains limited across population and service groups (Tables~\ref{tab:appendix_population_1} and~\ref{tab:appendix_population_2}). For visual impairment, JoyAI matches 6 of 46 intervention nodes (Table~\ref{tab:appendix_population_2}). Partial event coverage provides limited evidence that assistance will remain correct and selective throughout an entire video.

Correctly matched interventions can also be concentrated within a population. Thirteen of Gemini's nineteen correct matches belong to the adult group, corresponding to 13/229 for adults and 6/326 across the remaining groups. Its aggregate recall therefore conceals subgroup variation. Subgroup recall rates likewise require their denominators: the best scores for hearing and cognitive impairment represent only 6/17 and 2/9 nodes, respectively. The groups differ in task and observation conditions, and a single additional match would change these small-group rates. Practical selection should validate performance over entire videos for the intended users and tasks, including quiet periods; these comparisons do not isolate population effects.

\begin{table*}[!t]
\definecolor{scenarioGroup}{HTML}{F0F2F6}
\definecolor{scenarioDaily}{HTML}{91BFF0}
\definecolor{scenarioHealth}{HTML}{96C69A}
\definecolor{scenarioTravel}{HTML}{B49AD6}
\definecolor{scenarioEducation}{HTML}{FFB080}
\definecolor{scenarioSports}{HTML}{F5D967}

\centering
\caption{Full Correct (\%, higher is better, $\uparrow$) across five scenarios, evaluated on videos requiring intervention. Column headers report video counts. The highest value in each column is \textbf{bold}.}
\label{tab:scenario-full-correct}

\fontsize{9}{11}\selectfont
\setlength{\tabcolsep}{4pt}
\renewcommand{\arraystretch}{1.25}
\arrayrulecolor{black}

\begin{tabular*}{\textwidth}{
  @{\extracolsep{\fill}}lrrrrr@{}
}
\toprule

\textbf{Model}
& \multicolumn{1}{c}{
    \shortstack{
      \textbf{Daily living \&}\\
      \textbf{services} $\uparrow$\\
      95 videos\\[0.3ex]
      \textcolor{scenarioDaily}{\rule{1.25cm}{1.5pt}}
    }
  }
& \multicolumn{1}{c}{
    \shortstack{
      \textbf{Health \&}\\
      \textbf{caregiving} $\uparrow$\\
      70 videos\\[0.3ex]
      \textcolor{scenarioHealth}{\rule{1.25cm}{1.5pt}}
    }
  }
& \multicolumn{1}{c}{
    \shortstack{
      \textbf{Travel \&}\\
      \textbf{mobility} $\uparrow$\\
      40 videos\\[0.3ex]
      \textcolor{scenarioTravel}{\rule{1.25cm}{1.5pt}}
    }
  }
& \multicolumn{1}{c}{
    \shortstack{
      \textbf{Education \&}\\
      \textbf{work} $\uparrow$\\
      41 videos\\[0.3ex]
      \textcolor{scenarioEducation}{\rule{1.25cm}{1.5pt}}
    }
  }
& \multicolumn{1}{c}{
    \shortstack{
      \textbf{Sports \&}\\
      \textbf{leisure} $\uparrow$\\
      44 videos\\[0.3ex]
      \textcolor{scenarioSports}{\rule{1.25cm}{1.5pt}}
    }
  } \\
\midrule

\multicolumn{6}{c}{
  \cellcolor{scenarioGroup}\strut
  \textbf{Open-weight Models}
} \\

MiniCPM-o 4.5
& 1.1 & 0.0 & 2.5 & 2.4 & 2.3 \\

MOSS-VL
& \textbf{6.3} & \textbf{2.9} & 2.5 & 4.9 & \textbf{4.5} \\

MOSS Preview
& 0.0 & 0.0 & 0.0 & 0.0 & 0.0 \\

Venus-Omni
& 3.2 & 1.4 & 2.5 & 2.4 & 0.0 \\

Gander
& 3.2 & 1.4 & 0.0 & 4.9 & 0.0 \\

AURA
& 0.0 & 0.0 & 0.0 & 0.0 & 0.0 \\

JoyAI
& 0.0 & 0.0 & 0.0 & 0.0 & 0.0 \\

DuplexOmni (S1)
& 0.0 & \textbf{2.9} & 0.0 & 0.0 & 0.0 \\
\midrule

\multicolumn{6}{c}{
  \cellcolor{scenarioGroup}\strut
  \textbf{API-based Models}
} \\

Gemini 3.8 Live
& 5.3 & 0.0 & \textbf{5.0} & 4.9 & \textbf{4.5} \\

Qwen3.5 Plus
& 1.1 & 1.4 & 0.0 & 4.9 & \textbf{4.5} \\

Qwen3.5 Flash
& 2.1 & \textbf{2.9} & 2.5 & 4.9 & \textbf{4.5} \\

Qwen3.8 Flash
& 4.2 & 1.4 & 0.0 & \textbf{7.3} & 0.0 \\

\bottomrule
\end{tabular*}
\end{table*}

\paragraph{Scenario-level breakdown.}
Table~\ref{tab:scenario-full-correct} compares video-level success across five everyday scenarios. Mean Full Correct across the twelve configurations is highest in education and work (3.05\%), followed by daily living and services (2.19\%); the other three groups range from 1.19\% to 1.70\%. Eight configurations complete at least one video in each of these two groups. They are relatively more tractable for the tested configurations on the sampled tasks, although even their highest-scoring configurations complete only 3/41 and 6/95 videos, respectively.

The advantage is associated with the recorded task mix. Daily living and services has the largest proportion of single-intervention videos (73/95). In education and work, ten of the fifteen successful model-video evaluations come from two explicit-question videos, \texttt{ask\_school\_word} and \texttt{ask\_office\_tool}. Across all scenarios, all 65 fully correct model-video evaluations occur on single-intervention videos; none of the 97 videos requiring multiple interventions is fully correct for any configuration. Full Correct requires every intervention in a video to be handled correctly, making videos with more interventions harder to complete successfully. The scenario groups also differ in user populations and camera viewpoints, so these scores do not show that one scenario is inherently easier than another. Continuous-monitoring evaluation should check whether a model handles all required events and remains silent between them.

\paragraph{Intervention function and temporal demands.}
We further classify the 555 reference nodes into 68 preventive, 157 corrective or remedial, and 330 routine-support nodes using a model-assisted, internally reviewed procedure (Appendix~\ref{app:intervention_functions}). Prevention averts a specified problem, correction addresses an existing deviation, and routine support provides requested information, counts, answers, or state reports. The target problem determines the function rather than the action tag: in \texttt{elder\_box\_m\_pos}, warning of a coffee table ahead and providing post-fall help both carry \texttt{[warn]}, but serve different functions. Table~\ref{tab:intervention_function} groups the original main-experiment matches without rematching subsets; all required nodes remain in the denominators, including failed sessions.

\begin{table*}[!t]
\definecolor{functionGroup}{HTML}{E2E6ED}

\centering
\caption{Intervention Recall by primary service function
(\%, higher is better, $\uparrow$). Column headers give the number of required interval nodes. The highest value is \textbf{bold}.}
\label{tab:intervention_function}

\fontsize{9}{11}\selectfont
\setlength{\tabcolsep}{5pt}
\renewcommand{\arraystretch}{1.25}
\arrayrulecolor{black}

\begin{tabular*}{\textwidth}{
  @{\extracolsep{\fill}}lrrr@{}
}
\toprule

\textbf{ }
& \multicolumn{1}{c}{
    \shortstack{\textbf{Preventive} $\uparrow$\\68 nodes}
  }
& \multicolumn{1}{c}{
    \shortstack{\textbf{Corrective / remedial} $\uparrow$\\157 nodes}
  }
& \multicolumn{1}{c}{
    \shortstack{\textbf{Routine support} $\uparrow$\\330 nodes}
  } \\
\midrule

\multicolumn{4}{c}{
  \cellcolor{functionGroup}\strut
  \textbf{Open-weight Models}
} \\

JoyAI
& \textbf{30.88} & \textbf{26.75} & \textbf{14.55} \\

AURA
& 8.82
& 14.01
& 12.73 \\

MOSS-VL
& 1.47 & 7.01 & 6.06 \\

DuplexOmni (S1)
& 8.82 & 3.18 & 4.24 \\

Venus-Omni
& 0.00 & 1.91 & 2.42 \\

MiniCPM-o 4.5
& 0.00 & 0.64 & 2.12 \\

MOSS Preview
& 0.00 & 0.00 & 1.82 \\

Gander
& 0.00 & 0.64 & 1.52 \\
\midrule

\multicolumn{4}{c}{
  \cellcolor{functionGroup}\strut
  \textbf{API-based Models}
} \\

Qwen3.5 Flash
& 2.94 & 2.55 & 5.15 \\

Qwen3.5 Plus
& 4.41 & 1.91 & 4.55 \\

Gemini 3.8 Live
& 1.47 & 1.27 & 4.85 \\

Qwen3.8 Flash
& 4.41 & 1.27 & 3.03 \\
\bottomrule
\end{tabular*}
\end{table*}

Aggregate coverage can conceal a gap in the service a user needs. MOSS-VL recalls 7.01\% of corrective nodes and 6.06\% of routine-support nodes, but only one of 68 preventive nodes (1.47\%). Sixteen of Gemini's 19 correct matches are routine support, with only three across the 225 preventive and corrective nodes. Success in routine reporting therefore provides limited evidence of coverage when a task requires averting or managing a problem. Application validation should specify which service function must be delivered within its useful response window, rather than accepting aggregate recall as evidence of all three.

Prevention is not uniformly the least-covered function. JoyAI leads all three columns, with 30.88\% preventive, 26.75\% corrective, and 14.55\% routine recall. This coverage still coexists with low video-level reliability (Table~\ref{tab:omnilife-main}). A development target is therefore to preserve detection of actionable conditions while reducing unrelated output; blanket suppression or more frequent speech alone does not satisfy both requirements. The groups differ in task mix and observation conditions, so differences do not establish intrinsic function difficulty or identify an internal failure mechanism.

Temporal state is a separate requirement: 74 of the 400 tasks need information from previous observations, including 69 intervention videos. For example, the cup-counting task requires reporting the updated count and the final total (Figure~\ref{fig:failure_cases_compact}A). A correct final total does not compensate for missed intermediate updates, and the assistant must remain silent when no update is required. These tasks therefore require both accurate tracking and appropriately timed responses.

\subsection{Failure Taxonomy}
\label{sec:failure_taxonomy}

We manually categorize failures across all 4,800 main-experiment sessions, using recorded outputs, available audio transcripts, task references, event captions, and source frames where needed. The 23,337 output records support five observable failure categories (Table~\ref{tab:failure_cases}). Each category is counted at most once per completed session, and categories may co-occur. Rates use 4,674 completed sessions; 126 execution failures (2.63\% of all sessions) are reported separately and remain failures in the main evaluation. These categories summarize supported behavioral failures without establishing an exhaustive taxonomy or internal causes. Reference-dependent judgments exclude outputs beyond the annotation horizon. Appendix~\ref{app:failure_cases} presents four case studies with source frames, reference windows, response timelines, and descriptions of relevant outputs. Appendix~\ref{app:case_gallery} adds a 14-case gallery across user groups, all five primary scenarios, and the five failure categories.

\begin{table}[!t]
\centering
\caption{Observable failures across 4,674 completed sessions.
Categories may co-occur, and a flagged session may also contain
useful responses.}
\label{tab:failure_cases}

\normalsize
\setlength{\tabcolsep}{5pt}
\renewcommand{\arraystretch}{1.2}

\begin{tabularx}{\linewidth}{
    @{}
    >{\raggedright\arraybackslash}p{.24\linewidth}
    >{\raggedright\arraybackslash}X
    r
    @{}
}
\toprule
\textbf{Failure type} &
\textbf{Observable behavior} &
\textbf{Sessions (\%)} \\
\midrule

\textbf{F1:} Intervention control &
Misses required assistance, responds outside its window,
or speaks when no intervention is needed. &
3,223 (68.96) \\
\addlinespace[4pt]

\textbf{F2:} Situational grounding &
Misidentifies a current object, attribute, location,
relation, or risk state. &
148 (3.17) \\
\addlinespace[4pt]

\textbf{F3:} Temporal tracking &
Misreports accumulated counts, event order,
historical states, or completion status. &
219 (4.69) \\
\addlinespace[4pt]

\textbf{F4:} Task fulfillment &
Produces relevant speech but fails to complete
the requested assistance. &
450 (9.63) \\
\addlinespace[4pt]

\textbf{F5:} Unusable output &
Produces timestamps, empty tags, nonsemantic strings,
or an incomplete essential answer. &
1,064 (22.76) \\

\bottomrule
\end{tabularx}

\par\smallskip
\begin{minipage}{\linewidth}
\end{minipage}
\end{table}

\begin{figure}[!t]
\centering
\includegraphics[width=\linewidth]{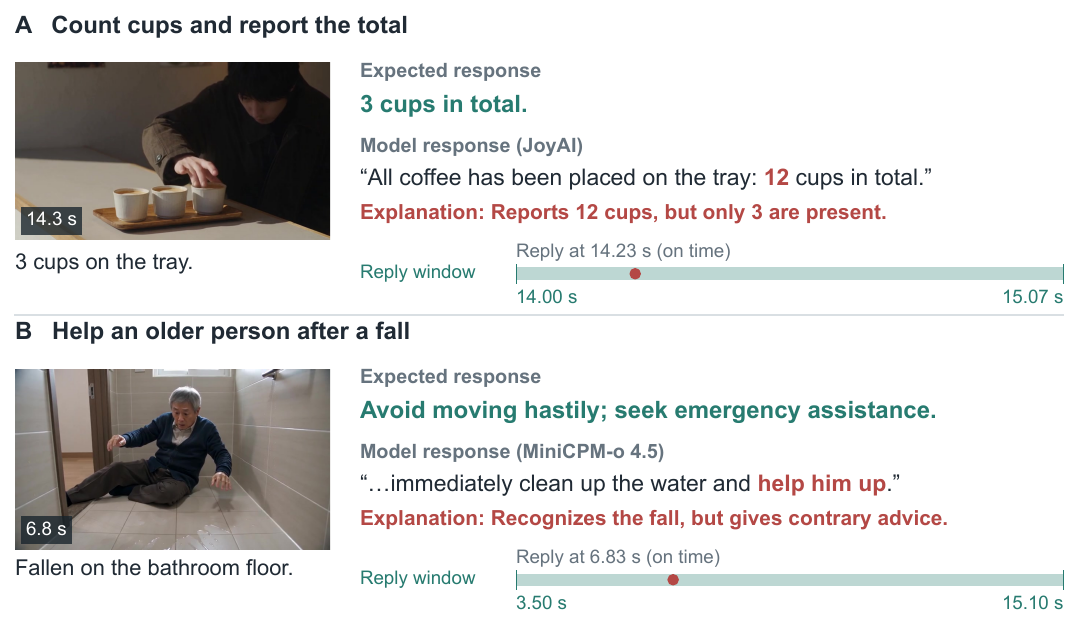}
\caption{Two on-time failures in cup counting (A) and post-fall assistance in a synthetic video (B), shown with source frames, condensed references, and translated outputs. Green intervals mark reference windows; red dots mark response onsets.}
\label{fig:failure_cases_compact}
\end{figure}

\paragraph{Intervention control is the most prevalent failure.}
Intervention-control failures occur in 68.96\% of completed sessions, including 1,720 intervention sessions with no recorded output. A model can also respond initially and then miss a later
event. For example, in a toy-car assembly video, DuplexOmni
comments at 0.72~s but gives no correction when an arm is
later installed incorrectly during the $[22,27)$~s window
(Appendix~\ref{app:failure_intervention}). Substantive output during reference silence is the converse control failure, even when its content is plausible. The development target is continued event monitoring with selective initiation and deadline control. Timing judgments concern recorded response onset, not the delivery time of every later word in a long response.

\paragraph{Current-scene grounding and temporal tracking require different checks.}
Situational grounding concerns what is present now: in \texttt{baby\_danger}, AURA describes the child as still eating and smiling at 14.70~s, while the child is choking and the mother is providing first aid (Appendix Figure~\ref{fig:case_gallery_1}). Temporal tracking concerns how observations accumulate: in \texttt{cnt\_cafe\_cup}, JoyAI repeatedly increments its count and finally reports twelve cups, although the completed sequence contains three (Figure~\ref{fig:failure_cases_compact}A). Verification of task-relevant entities and relations addresses the first observable error; event deduplication and explicit state updates are candidates for addressing the second. The trajectories distinguish these errors without identifying a memory mechanism.

\paragraph{Relevant speech can leave the requested assistance undone.}
Task-fulfillment failures occur in 450 sessions. In \texttt{grade\_slip\_bath\_pos}, a controlled synthetic bathroom-fall scenario, MiniCPM-o 4.5's sole response begins at 6.83~s, inside the $[3.5,15.1)$ assistance window (Figure~\ref{fig:failure_cases_compact}B). It identifies the fall but recommends immediately cleaning the water and lifting the person, contrary to the reference's request to avoid moving hastily and contact emergency assistance. The content judge marks this response incorrect. Recognizing the event and responding on time therefore do not establish task fulfillment: the proposed assistance must also satisfy the task's required response.

\paragraph{A recorded emission need not contain usable assistance.}
Unusable output occurs in 22.76\% of completed sessions. AURA records \texttt{<2.5 seconds>} at 3.07~s in \texttt{roof\_leak}, inside the $[2,5)$ leak-report window, without communicating the leak or its location. This category includes interface artifacts and incomplete essential answers, motivating separate protocol fields and checks of semantic completeness. Recorded artifacts do not establish what was audibly spoken or whether the model or interface produced them. Silence-control markers cannot satisfy a required response.

\begin{table*}[!t]
\definecolor{failureGroup}{HTML}{E2E6ED}

\centering
\caption{Model-wise failure prevalence (\%, lower is better, $\downarrow$) among completed sessions. Completed and Failed give session counts out of 400 attempts and the lowest value in each column is \textbf{bold}.}
\label{tab:failure_by_model}

\fontsize{9}{11}\selectfont
\setlength{\tabcolsep}{3pt}
\renewcommand{\arraystretch}{1.25}
\arrayrulecolor{black}

\begin{tabular*}{\textwidth}{
  @{\extracolsep{\fill}}lrrrrrrr@{}
}
\toprule
\textbf{Model}
& \textbf{Completed}
& \textbf{Control} $\downarrow$
& \textbf{Grounding} $\downarrow$
& \textbf{Tracking} $\downarrow$
& \textbf{Fulfillment} $\downarrow$
& \textbf{Unusable} $\downarrow$
& \textbf{Failed} $\downarrow$\\
\midrule

\multicolumn{8}{c}{
  \cellcolor{failureGroup}\strut
  \textbf{Open-weight Models}
} \\

JoyAI
& 400 & 97.50 & 16.50 & 19.75 & 47.75 & 56.50 & \textbf{0} \\

AURA
& 400 & \textbf{28.00} & 6.25 & 8.25 & 14.75 & 96.25 & \textbf{0} \\

MOSS-VL
& 400 & 66.25 & 1.50 & 4.00 & 7.50 & 1.75 & \textbf{0} \\

DuplexOmni (S1)
& 370 & 94.86 & 0.81 & 0.81 & 8.65 & \textbf{0.54} & 30 \\

Venus-Omni
& 400 & 67.50 & 0.50 & 2.75 & 6.75 & 5.50 & \textbf{0} \\

MiniCPM-o 4.5
& 400 & 67.50 & \textbf{0.25} & 1.25 & \textbf{1.75} & 3.25 & \textbf{0} \\

MOSS Preview
& 400 & 70.75 & 1.75 & 4.25 & 5.50 & 96.25 & \textbf{0} \\

Gander
& 400 & 67.50 & \textbf{0.25} & \textbf{0.75} & \textbf{1.75} & 0.75 & \textbf{0} \\
\midrule

\multicolumn{8}{c}{
  \cellcolor{failureGroup}\strut
  \textbf{API-based Models}
} \\

Qwen3.5 Flash
& 369 & 67.48 & 3.52 & 4.61 & 6.23 & 1.08 & 31 \\

Qwen3.5 Plus
& 369 & 67.48 & 2.98 & 4.34 & 5.42 & 1.36 & 31 \\

Gemini 3.8 Live
& 400 & 67.75 & 1.25 & 2.00 & 3.25 & 2.25 & \textbf{0} \\

Qwen3.8 Flash
& 366 & 66.39 & 2.19 & 3.01 & 5.19 & 0.82 & 34 \\
\bottomrule
\end{tabular*}
\end{table*}

\paragraph{Failure profiles differ across models.}
Table~\ref{tab:failure_by_model} distinguishes development priorities across systems. JoyAI combines a 97.50\% intervention-control failure rate with the highest prevalence of grounding, tracking, and task-fulfillment failures. AURA and MOSS Preview each contain unusable recorded output in 385 of 400 sessions (96.25\%); these sessions can also contain useful responses. For the three Qwen configurations, intervention-control failures occur in 66.39\% to 67.48\% of completed sessions, while the other types are less prevalent. The profiles motivate testing output validation and task completion alongside selective initiation, according to each system's behavior.

\paragraph{Low content-error prevalence must be interpreted alongside response coverage.}
Gander and MiniCPM each exhibit grounding errors in only 0.25\% of completed sessions, but emit nothing in 260 and 245 of their 290 completed intervention sessions, respectively. Gemini emits nothing in 230 of 290 such sessions. Few observed content errors can therefore coexist with extensive unmet assistance. These rates measure prevalence across sessions; accuracy conditional on producing a response has a different denominator. They should be read with Intervention Recall and Silence Success, while execution failures remain in the main evaluation.

\subsection{Ablation Studies}
\label{sec:ablations}
We test whether changing or repeating the task instruction, delivering it through speech, or allowing later responses can reduce missed interventions.
\paragraph{Input design and periodic reminders.}
We compare five text-input conditions on the same 400 videos for MiniCPM, MOSS-VL, Venus, Gemini, and Qwen3.5 Flash: the original task text, a structured instruction containing the same task information, repeating the full task every second, repeating it every 5 seconds, and sending a general reminder every second to continue the original task and remain silent unless its response condition is met. Source media, model settings, and content scoring remain fixed within each model; no condition uses task TTS. Reminders use each model's native text interface, retaining its acceptance or rejection behavior without forcing generation. Table~\ref{tab:input_design_ablation} compares timely coverage and unwanted responses during reference silence.

\begin{table*}[!t]
\definecolor{ablationTeal}{HTML}{008577}
\definecolor{ablationRed}{HTML}{C43E2B}

\definecolor{ablationTealBg}{HTML}{E5F3EF}
\definecolor{ablationRedBg}{HTML}{F7EBE8}

\centering
\caption{Input design and periodic task reminders.
Upper: required events handled correctly and on time
(\%, higher is better, $\uparrow$). Lower: unwanted responses per minute of reference silence (lower is better, $\downarrow$). Within each model, the highest values in the upper section
are shown in \textcolor{ablationTeal}{\textbf{bold}}, and the lowest values in the lower section are shown in \textcolor{ablationRed}{\textbf{bold}}.}
\label{tab:input_design_ablation}

\fontsize{9}{11}\selectfont
\setlength{\tabcolsep}{3pt}
\renewcommand{\arraystretch}{1.3}
\arrayrulecolor{black}

\begin{tabularx}{\textwidth}{
  @{}
  >{\raggedright\arraybackslash}p{0.30\textwidth}
  *{5}{>{\centering\arraybackslash}X}
  @{}
}
\toprule

\textbf{Input strategy}
& \textbf{MiniCPM-o 4.5}
& \textbf{MOSS-VL}
& \textbf{Venus-Omni}
& \textbf{Gemini 3.8 Live}
& \textbf{Qwen3.5 Flash} \\
\midrule

\multicolumn{6}{c}{
  \cellcolor{ablationTealBg}\strut
  \textcolor{ablationTeal}{
    \textbf{Required events handled correctly and on time (\%)}
    $\uparrow$
  }
} \\
\midrule

Original task text
& 1.44 & 5.77 & 1.98 & 3.42
& \textcolor{ablationTeal}{\textbf{4.14}} \\

Structured task text
& 1.44 & 6.31 & 4.32 & 2.52 & 3.60 \\

Repeat full task every second
& \textcolor{ablationTeal}{\textbf{5.41}}
& 6.85
& \textcolor{ablationTeal}{\textbf{13.51}}
& 12.43
& 3.42 \\

Repeat full task every 5 seconds
& 1.80
& \textcolor{ablationTeal}{\textbf{14.05}}
& 10.27
& \textcolor{ablationTeal}{\textbf{20.54}}
& 3.42 \\

General reminder every second
& 2.88 & 3.78 & 6.67 & 11.71 & 3.24 \\
\midrule

\multicolumn{6}{c}{
  \cellcolor{ablationRedBg}\strut
  \textcolor{ablationRed}{
    \textbf{Unwanted responses per minute of reference silence}
    $\downarrow$
  }
} \\
\midrule

Original task text
& \textcolor{ablationRed}{\textbf{0.14}}
& 1.41
& \textcolor{ablationRed}{\textbf{0.48}}
& \textcolor{ablationRed}{\textbf{0.57}}
& 1.34 \\

Structured task text
& 0.18
& \textcolor{ablationRed}{\textbf{1.20}}
& 0.60
& 0.64
& 1.34 \\

Repeat full task every second
& 0.64 & 5.13 & 21.68 & 4.96 & 1.29 \\

Repeat full task every 5 seconds
& 0.20 & 4.48 & 7.58 & 5.16 & 1.29 \\

General reminder every second
& 0.43
& 1.73
& 10.72
& 4.06
& \textcolor{ablationRed}{\textbf{1.25}} \\
\bottomrule
\end{tabularx}
\end{table*}

Structuring the task produces small, mixed changes in recall, while periodic reminders yield larger gains for four models. The better tested reminder interval differs across models, and no variant improves Qwen's recall over the original task. Reminders both repeat task information and send new messages that may trigger responses. We compare full-task and general reminders, both sent every second, to examine whether repeating
the task details helps. Full-task reminders yield higher recall for all five models, although Qwen gains only one additional match. This suggests that reminder content matters, but does not show that the models had forgotten the task. Message length and interface behavior may also affect the results.

Higher intervention coverage can come with more unwanted output.
Gemini's 5-second reminders raise recall from 3.42\% to 20.54\%,
while increasing unwanted responses from 0.57 to 5.16 per minute
of reference silence. Venus's 1-second schedule reaches 13.51\%
recall while producing 21.68 unwanted responses per minute of
reference silence. For applications requiring sustained monitoring,
reminder schedules should therefore be evaluated on separate
validation videos containing both required interventions and
quiet intervals. The tested frequencies provide candidate settings
for validation rather than a universal deployment default.

A concrete interface direction is persistent task state containing the event condition, required assistance, and whether the event has already been handled, with updates separated from response generation. Where supported, such updates could be compared with ordinary repeated messages under matched task content, media timeline, and update frequency. Retaining coverage while reducing unwanted output would support this separation. The present ablation motivates that test but does not validate the interface.

\paragraph{Task instruction modality.}
We compare written and spoken delivery of the same task for MiniCPM-o 4.5, MOSS-VL, Venus-Omni, and Qwen3.5 Flash, using 400 videos per model. The spoken condition removes the case-specific task text from the initial instruction and plays that task once before the video; the general benchmark instruction, source media, model settings, and content judge remain fixed. MOSS-VL receives the actual SenseVoice transcript without a task-text fallback. Scoring uses the original video timeline: responses before video onset are retained separately, and responses crossing that boundary are scored on their video-phase portion. All cases remain in the denominators, including Venus's 30 native input failures and Qwen's three unsuccessful spoken-task sessions.

\begin{figure}[!t]
    \centering
    \includegraphics[width=\linewidth]{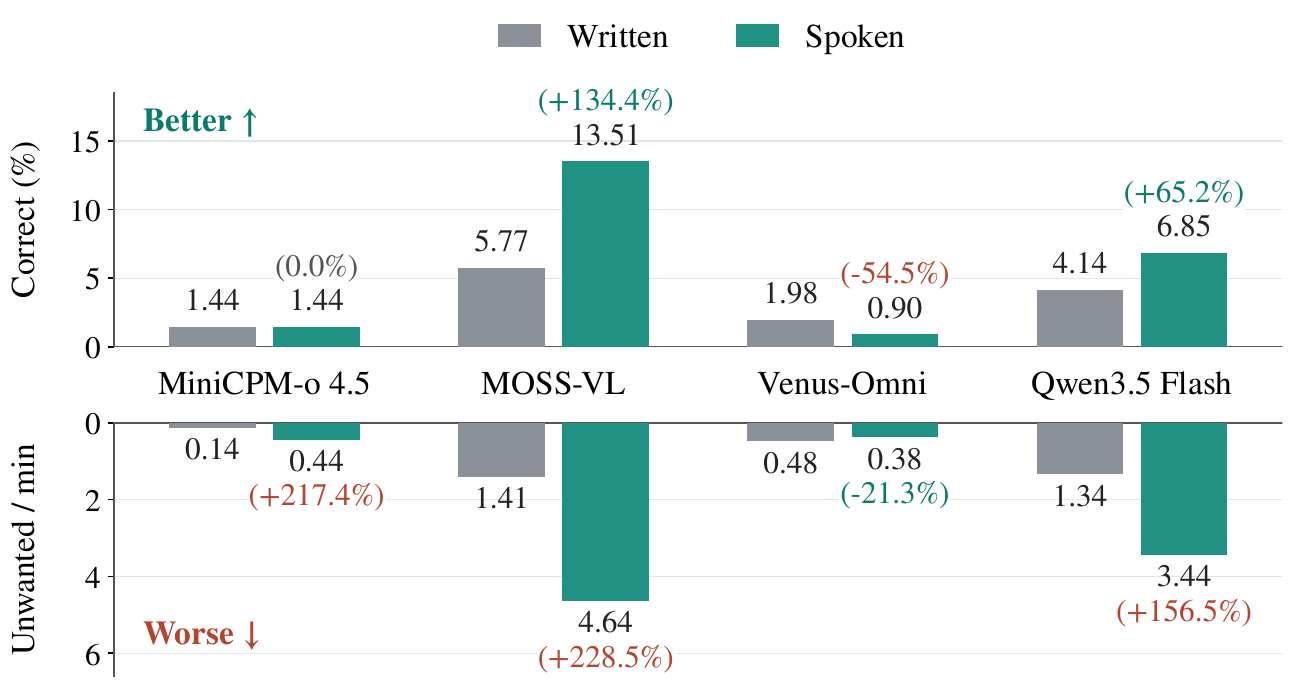}
    \caption{Effect of task instruction modality on Intervention Recall (top; higher is better) and unwanted responses per minute of reference silence (bottom; lower is better).}
    \label{fig:instruction_modality}
\end{figure}

In Figure~\ref{fig:instruction_modality}, the lower panel shows unwanted responses per minute with positive rate labels; percentages in parentheses are relative changes computed from unrounded scores. Both panels retain all 400 videos, including failed sessions; Correct uses the 555 required nodes, while Unwanted uses the total duration of reference silence.

Changing task delivery does not consistently improve
performance (Figure~\ref{fig:instruction_modality}).
MiniCPM's response count during the video increases from
54 to 129, but its correct, timely matches remain at eight. MOSS-VL and Qwen gain recall but have lower Full Correct scores (Table~\ref{tab:appendix_instruction_modality}), whereas Venus gains no recall. More responses therefore
do not necessarily improve performance over an entire video.

However, this comparison changes more than the task modality. It also changes how and when the task reaches the model, which may affect when the model starts responding.

For MOSS-VL, the task changes from an initial system instruction to one or more ASR-derived user prompts. In the deployed implementation, each new prompt opens an assistant turn. Raw input logs show that the final task segment arrives after video onset in 399 of 400 cases, with a median delay of 0.26~s. ASR segmentation and arrival timing can thus initiate or restart responses during observation, even though the model receives text. Intervention videos with no output fall from 194 to 37, yet the share of emitted responses matched correctly and on time falls from 9.12\% to 7.22\%. The extra output also extends beyond task intake: unwanted responses beginning at or after 10~s increase from 202 to 378. For example, in a reference-silent night-duty video, MOSS continues describing the scene at 12.68~s while stating that no abnormality has been found. Observations support an interaction-level explanation involving response activation and weak selectivity. They do not isolate the contributions of transcription, task placement, and segmentation.

Qwen's net increase is more concentrated at startup. Among the 366 videos with source audio completed in both conditions, unwanted replies beginning in the first 2~s increase from 1 to 333, while later unwanted replies decrease from 198 to 183. Correct matches in the early interval increase from 0 to 17, while later matches decrease from 17 to 13. An initial response can therefore recover an event already present at onset or break required silence when no intervention is needed. In a basketball-question video, Spoken replies at 0.61~s, ``Okay, I am watching. Which basketball rule would you like to know?'' (translated), although the reference requires silence until 11.5~s. This pattern is consistent with a conversational response to task intake.

A controlled follow-up should compare the original task text and actual ASR transcript under matched message roles and arrival times, then compare complete and segmented delivery. The timing buckets are diagnostic: suppressing startup responses could also remove legitimate interventions already required at video onset.

\paragraph{Relaxing the response deadline.}
To assess how much intervention failure can be recovered by allowing later responses, we extend each reference window from $[s,e)$ to $[s,e+\delta)$, with $\delta\in\{0,1,2,5,10\}$ seconds. Window starts, content judgments, and recorded model outputs remain fixed, and we recompute maximum-cardinality, order-preserving one-to-one matches at each tolerance. Intervention Recall retains all 555 required nodes per model, including failed sessions. Gemini and Qwen use the text-prompt main-experiment runs without task TTS. This offline scoring ablation adds no inference and does not extend output collection.

\begin{figure}[!t]
    \centering
    \includegraphics[width=\linewidth]{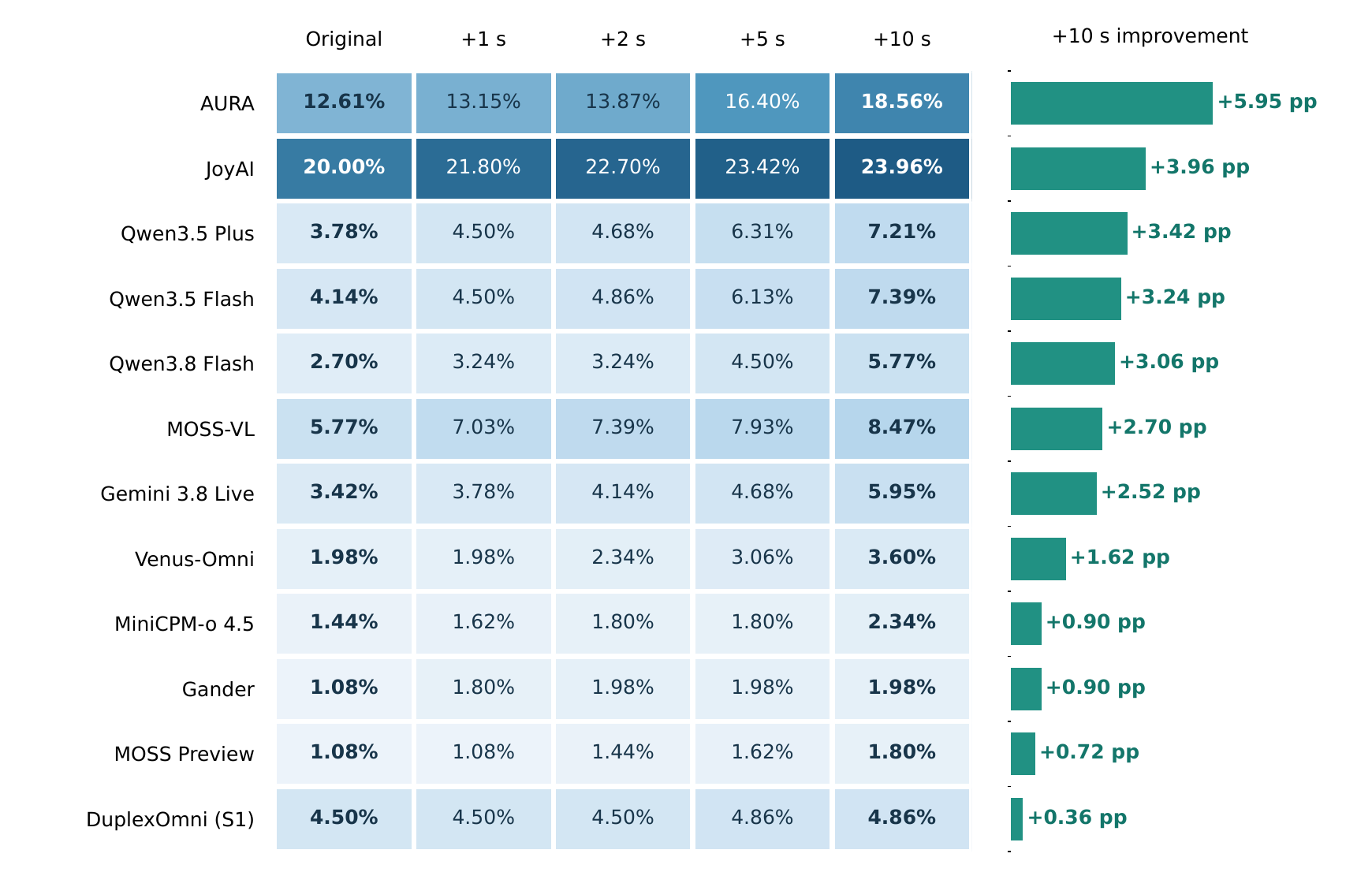}
    \caption{Intervention Recall (\%) as reference-window ends are extended. Rows are ordered by the gain at $\delta=10$~s, shown in percentage points (pp) by the right-hand bars.}
    \label{fig:tolerance_ablation}
\end{figure}

\paragraph{How much failure does extra time recover?}
Figure~\ref{fig:tolerance_ablation} measures how much usable late content is already present in the recorded outputs. Extending deadlines by 10~s increases recall by 0.36--5.95 percentage points. AURA gains the most but reaches only 18.56\%, and all models remain below 24\%. As a fraction of the nodes missed under the original windows, the recovered share $(TP_{10}-TP_0)/(555-TP_0)$ ranges from 0.38\% to 6.80\%; at least 93.20\% remain unmatched. Thus the late-correct replies account for a small part of the deficit under the original deadlines.

This is deadline sensitivity of fixed outputs, not an estimate of the benefit of faster inference. Extended windows may overlap later events, and a compatible late reply need not have been intended for an earlier reference. The original collection horizon also limits observable lateness. Applications should set deadlines according to when assistance remains useful and test delivery improvements on late-correct cases, with timestamps across input, buffering, generation, and receipt. Absent and incorrect responses require separate tests of initiation and task fulfillment.

\section{Conclusion}
\label{sec:conclusion}

Omni-Life evaluates whether real-time omnimodal assistants provide the required help at the right moment while remaining silent when no intervention is needed. Across 400 videos with 555 required intervention nodes, none of the twelve main-experiment configurations reliably combines event coverage, useful content, and appropriate silence; the strongest Full Correct score is 4.48\% among the twelve evaluated model configurations. The ablations expose a practical distinction between activating responses and improving sustained assistance. Reminder and task-delivery changes can recover required events while also increasing unwanted output, and extending response deadlines by 10~s recovers at most 6.80\% of originally unmatched nodes within the recorded trajectories. These findings motivate evaluation and model development that jointly address persistent task execution, selective response initiation, and timely task fulfillment.
\section{Limitations}
\label{sec:limitations}
Omni-Life's finite recordings, predefined tasks, and response windows limit generalization to prolonged deployment, adaptive interaction, or changing preferences. Synthetic clips may favor explicit events, and the case mix may not represent everyday support needs. Population and scenario comparisons depend on tasks and observation conditions, establishing neither causal population effects nor equitable performance. System comparisons depend on interfaces and input pipelines; onset measures client-side receipt, not audible playback. Reference-only automatic judging may miss distinctions in task fulfillment. The four-annotator baseline may not generalize broadly, and QA spot checks do not validate all annotations or judgments. Deadline relaxation rescores fixed outputs without testing faster models or longer recordings. Separating task updates from response generation requires further validation of assistance quality and unwanted output.
\section{Ethics Statement}
\label{sec:ethics}

The experiments evaluate recorded video streams rather than intervening in live situations. Synthetic generation permits hazardous situations to be represented without deliberate exposure of participants to those hazards. Scenarios involving children, older adults, and people with disabilities describe contextual support needs; they should not be interpreted as diagnoses, fixed deficits, or assumptions about every member of a group. Public accessibility of real-world clips does not establish consent or permission for redistribution, and dissemination requires source-specific rights and privacy review. Benchmark performance is evidence about model behavior under the protocol and does not establish readiness to replace human oversight in consequential assistance.


\bibliography{refs}
\bibliographystyle{omni-life}

\clearpage
\clearpage
\appendix
\input{appendices/contents}
\section{Dataset Records and Evaluation Details}
\label{app:data_construction}
\begin{figure}[p]
\centering
\includegraphics[width=\linewidth]{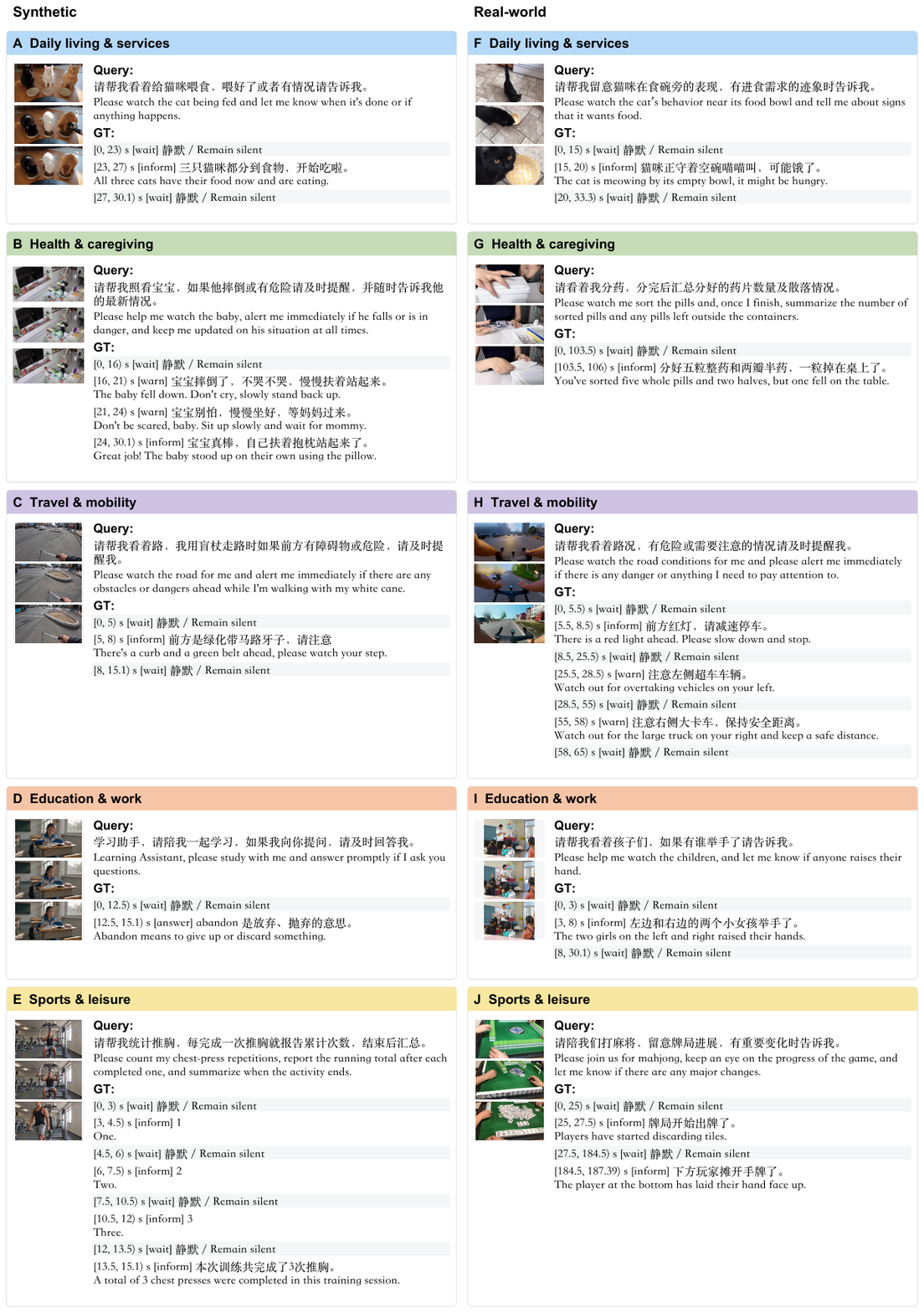}
\caption{Ten dataset examples across five scenario domains,
pairing synthetic (left) and real-world (right) videos.
Each case shows three source frames in temporal order,
bilingual task queries, reference responses, and intervals
of required assistance and silence.}
\label{fig:dataset_examples}
\end{figure}
\subsection{Evaluation Set and Annotation Structure}
Evaluation records correspond to videos rather than the life stories used for scenario construction; Table~\ref{tab:dataset_statistics} reports their source composition and annotation counts. All main-experiment configurations use the same frozen records.
\paragraph{Record structure.} Each record contains a case identifier, a standing task, contextual metadata, and a sequence of annotations. An annotation specifies a half-open source-time interval $[s,e)$, an action, and Chinese and English reference content where a response is required. The two reference languages express the same requirement; a model need not produce both.
\paragraph{Reference actions.} The four actions requiring a response are \texttt{[inform]}, \texttt{[warn]}, \texttt{[feedback]}, and \texttt{[answer]}. The fifth category, \texttt{[wait]}, requires no response. Counting and summarization requirements are included under \texttt{[inform]} in this release. Task-specific information, such as the correct count or completed state, is required for content correctness.
\paragraph{Dataset examples.}
Figure~\ref{fig:dataset_examples} presents ten examples
across the five scenario domains, pairing synthetic and
real-world videos. Each example shows source frames, bilingual task queries, reference responses, and intervals
of required assistance and silence.

\subsection{Construction and Quality-Control Criteria}
Scenario construction starts from a support need and defines an observable condition under which a particular response becomes useful. This condition depends on the standing task, rather than every visible change in the scene. For example, observing an object does not necessarily justify describing it when the task requires waiting for a question, a completed stage, or a hazard. A reference window specifies the interval in which the assistance is appropriate, rather than the first moment at which an object can be recognized.

The construction criteria distinguish evidence available in the clip from information used only for evaluation. Decision-relevant objects, actions, and state changes must be observable; an intended scenario alone cannot justify an annotation.
Required answers must not be supplied as narration, subtitles, or other answer leakage. Synthetic clips additionally require plausible event progression and consistent participants and objects.
Non-intervention intervals are explicit requirements, including in paired cases where the task-relevant condition is absent. These criteria describe the intended inclusion and annotation rules; they do not establish that every possible ambiguity has been eliminated or replace an independent annotation-reliability study.

The recorded tasks, media identities, and annotations are frozen before scoring. Each model receives the standing task and source media, while reference content, annotation windows, event captions, and QA answer keys are withheld from streaming inference.
Population and scenario metadata support descriptive analyses of the recorded case mix. A task-derived assistant role does not establish which physical device recorded a video or hosted an assistant, and is not used as a substitute for a verified device label.

\subsection{Inference and Recording Configuration}
\label{app:inference_configuration}

\paragraph{Media delivery and session completion.} Streaming inputs retain each interface's native chunking: 1~s for MiniCPM, Gander, and Venus, and 480~ms for DuplexOmni. Gemini and Qwen receive 200-ms audio packets and 640-pixel-wide frames.

After source EOF, local adapters finish pending processing without adding media; MOSS additionally collects output for 2~s after its ASR flush. Gemini and Qwen collect output for 2~s after clip end. The ASR front ends transcribe the original source audio, rather than synthesizing speech from the task prompt.

\paragraph{Streaming generation.} MiniCPM uses greedy decoding with at most 20 speech-text tokens per step. Gander samples with temperature $T=0.7$, top-$p=0.8$, top-$k=20$, and an 11-token limit; Venus uses $T=0.7$, top-$p=0.8$, top-$k=100$, a 50-token limit, and length penalty 0.8.
AURA uses $T=0.5$ and a 128-token limit; JoyAI uses $T=0.8$, top-$p=0.9$, top-$k=40$, and a 128-token limit. DuplexOmni (S1) uses $T=0.01$, top-$p=1$, and a 999-token limit. These limits apply to native generation steps, not the entire video.
Both MOSS models decode greedily using native silence and stop controls rather than an active token-length cap; MOSS-VL is paced at 12 tokens/s, while Preview uses its effectively unthrottled default.
Gemini and Qwen use $T=0$: Gemini enables proactive audio and sliding-window context management, while Qwen uses semantic voice-activity detection and seed 0.

\paragraph{Output normalization and records.} Gemini's native audio transcription is supplemented by ASR when a nonempty audio response lacks text. Response events preserve text, an optional tag, and source-aligned start/end timestamps. Per-case records preserve inputs, raw and parsed outputs, latency, runtime settings, failures, and reported API usage. Control markers are normalized; nonempty invalid responses remain false-positive candidates.

Local systems and Qwen return text, while Gemini's onset is the first audio packet; onset otherwise denotes the first received text content. These intervals measure client-side delivery rather than loudspeaker playback.

For DuplexOmni (S1), the adapter groups consecutive nonempty \texttt{tts} fields from complete HTTP responses into an observable response segment, closed by an empty field, a \texttt{[STOP]} control, or source EOF. Normal native end-of-turn boundaries are not exposed by this interface.

\paragraph{QA generation.} Local models use greedy decoding with a 64-token limit. Qwen and Gemini use temperature 0 with token limits of 64 and 1,024, respectively.

\subsection{Content Judgment and One-to-One Matching}
\label{app:content_matching}
DeepSeek-V4-Flash-0731 uses temperature 0, disabled thinking, and a 32-token limit. For an active reference node $i$ and recorded response $j$, the binary content judgment $c_{ij}$ assesses whether the response conveys the minimum information required by that node's task and reference, without a substantive contradiction. Semantically equivalent wording and concise answers are accepted. Missing essential information, an incorrect entity or state, a monitoring promise in place of a required event report, and an emission without a usable answer are rejected. Optional reference details and behavior-tag correctness are not requirements for content credit.

Let $t_j$ be the recorded onset of response $j$. An eligible main-score edge satisfies
\begin{equation}
c_{ij}=1 \quad\text{and}\quad s_i \leq t_j < e_i.
\end{equation}
We select a maximum-cardinality matching whose node and response indices both increase. This preserves temporal order and allows each node and each response to contribute at most one match. Responses are not reassigned using a guessed intended event, a timestamp mentioned in their text, or a corrected behavior tag.

Reference-window ends are extended only in the explicitly reported tolerance analysis. The content judge is held fixed when recomputing those matches.

\subsection{Silence, Video-Level Correctness, and Failures}
\label{app:silence_metrics}
The unwanted-response rate counts recorded response onsets inside reference \texttt{[wait]} intervals, divided by the total duration of those intervals in minutes. It includes silence intervals within intervention videos as well as fully silent videos. Thus it is distinct from the false-positive count in Intervention F1, which includes every unmatched response, including a wrong answer during an active window. Neither measure treats frequent output as assistance by itself.

Full Correct applies to the 290 intervention videos. It requires every active node to be matched, no extra response, no recorded response interval overlapping a silence interval, and successful session completion. Its interval-overlap check is stricter than counting only unwanted response onsets: a response can begin inside an active window yet continue into a quiet interval.

Silence Success instead applies to the 110 fully silent videos and requires a completed session with no actual output. Empty bookkeeping records and recognized control markers do not create useful assistance; a recorded nonempty response cannot be reclassified as successful silence merely because its behavior tag is missing or invalid.

All planned cases remain in the applicable denominators. An execution failure is reported separately from a completed session that emits no response, and cannot receive Full Correct or Silence Success credit. Interface-specific response grouping and source-aligned timestamps are retained.

\subsection{Human Baseline}
\label{app:human_baseline}

Four annotators independently completed Chinese QA and streaming evaluations. Scores are unweighted annotator means, rounded after averaging, and measure reference-based performance rather than annotator agreement. Each annotator's performance was first calculated separately, preserving the same evaluation units and scoring criteria across annotators.

\paragraph{QA scoring.}
Each annotator answers 400 content and 400 timing questions under the same scoring protocol, with missing or invalid answers counted as incorrect. Content and Timing Accuracy each use a denominator of 400; Joint Accuracy requires both answers per video to be correct across all 400 videos. Thus, a video contributes to Joint Accuracy only when both its content answer and its timing answer are correct. Correctness on either question alone is insufficient, and Joint Accuracy is not the arithmetic mean of the two component accuracies.

\paragraph{Streaming scoring.}
A judge assessed whether each response conveyed the essential information required by the task and the corresponding reference intervention. Semantically equivalent wording was accepted; missing essential information or substantive contradictions were considered incorrect. Human responses used the same timing rules, matching procedure, and metric definitions as model responses. A response received matching credit only when its content was correct and its recorded onset fell within the original reference interval $[s_i,e_i)$. Early or late responses received no matching credit. Content correctness and temporal validity were therefore jointly required: neither a timely but incorrect response nor a correct response outside the interval qualified. Matching was one-to-one, so each response and reference intervention could contribute at most one correct match.

\paragraph{Score aggregation.}
For each metric, the reported human baseline is the unweighted arithmetic mean of the four annotators' unrounded scores on the same benchmark. Rounding is applied only to the final mean. Each annotator therefore contributes equally to the reported baseline under the shared evaluation protocol. These scores measure performance against the benchmark references rather than agreement between annotators.

\section{Complete Streaming Results}
\label{app:complete_streaming_results}

The following figures and tables supplement the main results using the same frozen content judgments and timing rules. Unless stated otherwise, every model or condition retains 400 videos, 555 required intervention nodes, 290 intervention videos, and 110 fully silent videos, including unsuccessful sessions. Intervention Recall and F1, Full Correct, and Silence Success follow the evaluation protocol in the main text. Unwanted responses are counted during annotated silence and divided by its duration in minutes. All displayed rates are rounded after aggregation.

\subsection{Population, Scenario, and Reference-Action Breakdown}

Tables~\ref{tab:appendix_population_1} and~\ref{tab:appendix_population_2} report all population and service-group categories, with video and intervention-node counts. These are descriptive comparisons of the recorded tasks. Groups differ in scenario, observation conditions, and required interventions, and the smallest groups contain only nine required nodes. The results do not isolate a causal population effect or estimate performance for an entire real-world population. Pets and multi-person settings are retained as service groups in the original case labels.

\definecolor{rawtitle}{HTML}{E5F1EF}
\begin{table}[!t]
\centering
\caption{Population and service-group Intervention Recall (\%), part 1 of 2. Cases and required nodes are fixed across models; failed sessions remain in the node denominators.}
\label{tab:appendix_population_1}
\begingroup
\small
\setlength{\tabcolsep}{4pt}
\setlength{\aboverulesep}{0pt}
\setlength{\belowrulesep}{0pt}
\setlength{\extrarowheight}{2pt}
\newcommand{\HT}[2]{%
  \begin{tabular}[c]{@{}c@{}}\textbf{#1}\\\textbf{#2}\end{tabular}%
}
\resizebox{\linewidth}{!}{%
\begin{tabular}{@{}lcccccccc@{}}
\arrayrulecolor{black}\toprule[0.8pt]
\rowcolor{rawtitle}
\textbf{User / service group} &
\textbf{Cases} &
\textbf{Nodes} &
\HT{MiniCPM-o}{4.5} &
\textbf{MOSS-VL} &
\HT{MOSS}{Preview} &
\HT{Venus-}{Omni} &
\textbf{Gander} &
\textbf{AURA} \\
\midrule
\rowcolor{black!3}
Adults & 172 & 229 & 3.06 & 4.37 & 0.44 & 1.75 & 1.75 & 14.41 \\
Infants and toddlers & 26 & 34 & 0.00 & 0.00 & 0.00 & 0.00 & 0.00 & 2.94 \\
\rowcolor{black!3}
Schoolchildren & 36 & 31 & 0.00 & 3.23 & 0.00 & 3.23 & 3.23 & 9.68 \\
Older adults & 29 & 58 & 1.72 & 8.62 & 1.72 & 1.72 & 0.00 & 22.41 \\
\rowcolor{black!3}
Visual impairment & 27 & 46 & 0.00 & 2.17 & 4.35 & 0.00 & 0.00 & 2.17 \\
Hearing impairment & 13 & 17 & 0.00 & 11.76 & 0.00 & 5.88 & 0.00 & 17.65 \\
\rowcolor{black!3}
Color-vision deficiency & 11 & 12 & 0.00 & 0.00 & 0.00 & 0.00 & 0.00 & 16.67 \\
Mobility impairment & 8 & 11 & 0.00 & 0.00 & 0.00 & 0.00 & 0.00 & 0.00 \\
\rowcolor{black!3}
Cognitive impairment & 11 & 9 & 0.00 & 0.00 & 0.00 & 11.11 & 0.00 & 0.00 \\
Chronic conditions & 13 & 9 & 0.00 & 0.00 & 0.00 & 0.00 & 0.00 & 11.11 \\
\rowcolor{black!3}
Multi-person settings & 24 & 60 & 0.00 & 18.33 & 1.67 & 3.33 & 0.00 & 13.33 \\
Pets & 30 & 39 & 0.00 & 5.13 & 2.56 & 2.56 & 2.56 & 12.82 \\
\arrayrulecolor{black}\bottomrule[0.8pt]
\end{tabular}%
}
\endgroup
\end{table}
\definecolor{rawtitle}{HTML}{E5F1EF}
\begin{table}[!t]
\centering
\caption{Population and service-group Intervention Recall (\%), part 2 of 2. Cases and required nodes are fixed across models; failed sessions remain in the node denominators.}
\label{tab:appendix_population_2}
\begingroup
\normalsize
\setlength{\tabcolsep}{3pt}
\setlength{\aboverulesep}{0pt}
\setlength{\belowrulesep}{0pt}
\setlength{\extrarowheight}{1pt}
\renewcommand{\arraystretch}{1.15}
\sbox0{\bfseries DuplexOmni}\edef\wDup{\the\wd0}%
\sbox0{\bfseries Qwen3.5}\edef\wQw{\the\wd0}%
\sbox0{\bfseries Gemini}\edef\wGem{\the\wd0}%
\sbox0{\bfseries 3.8 Live}\ifdim\wd0>\wGem \edef\wGem{\the\wd0}\fi
\sbox0{\bfseries JoyAI}\edef\wJoy{\the\wd0}%
\sbox0{\bfseries Cases}\edef\wCases{\the\wd0}%
\sbox0{\bfseries Nodes}\edef\wNodes{\the\wd0}%
\newcommand{\pophead}[2]{%
  \multicolumn{1}{>{\columncolor{rawtitle}}c}{%
    \parbox[c][2.7em][c]{#1}{\centering\bfseries #2}%
  }%
}
\begin{tabularx}{\linewidth}{@{}>{\raggedright\arraybackslash}Xcccccccc@{}}
\arrayrulecolor{black}\toprule[0.8pt]
\multicolumn{1}{>{\columncolor{rawtitle}}X}{%
  \parbox[c][2.7em][c]{\hsize}{\centering\bfseries User / service group}%
} &
\pophead{\wCases}{Cases} &
\pophead{\wNodes}{Nodes} &
\pophead{\wJoy}{JoyAI} &
\pophead{\wDup}{DuplexOmni\\(S1)} &
\pophead{\wGem}{Gemini\\3.8 Live} &
\pophead{\wQw}{Qwen3.5\\Plus} &
\pophead{\wQw}{Qwen3.5\\Flash} &
\pophead{\wQw}{Qwen3.8\\Flash} \\
\midrule
\rowcolor{black!3}
Adults & 172 & 229 & 21.40 & 3.06 & 5.68 & 3.49 & 4.37 & 3.49 \\
Infants and toddlers & 26 & 34 & 8.82 & 0.00 & 0.00 & 2.94 & 2.94 & 2.94 \\
\rowcolor{black!3}
Schoolchildren & 36 & 31 & 35.48 & 3.23 & 3.23 & 3.23 & 3.23 & 3.23 \\
Older adults & 29 & 58 & 27.59 & 6.90 & 0.00 & 0.00 & 0.00 & 0.00 \\
\rowcolor{black!3}
Visual impairment & 27 & 46 & 13.04 & 6.52 & 0.00 & 6.52 & 2.17 & 4.35 \\
Hearing impairment & 13 & 17 & 35.29 & 0.00 & 0.00 & 0.00 & 5.88 & 5.88 \\
\rowcolor{black!3}
Color-vision deficiency & 11 & 12 & 41.67 & 8.33 & 0.00 & 0.00 & 0.00 & 0.00 \\
Mobility impairment & 8 & 11 & 18.18 & 0.00 & 0.00 & 0.00 & 0.00 & 0.00 \\
\rowcolor{black!3}
Cognitive impairment & 11 & 9 & 22.22 & 0.00 & 0.00 & 0.00 & 0.00 & 0.00 \\
Chronic conditions & 13 & 9 & 0.00 & 0.00 & 0.00 & 0.00 & 0.00 & 0.00 \\
\rowcolor{black!3}
Multi-person settings & 24 & 60 & 8.33 & 10.00 & 5.00 & 10.00 & 11.67 & 1.67 \\
Pets & 30 & 39 & 15.38 & 7.69 & 5.13 & 5.13 & 5.13 & 2.56 \\
\arrayrulecolor{black}\bottomrule[0.8pt]
\end{tabularx}
\endgroup
\end{table}

Table~\ref{tab:scenario_results} provides the complementary Intervention Recall breakdown for the five scenarios. These groups partition the 555 required intervention nodes. Recall measures the fraction of correctly matched nodes, while Full Correct in Table~\ref{tab:scenario-full-correct} requires the entire intervention video to succeed without extra output or silence overlap.

\definecolor{rawtitle}{HTML}{E5F1EF}

\begin{table}[!t]
\centering
\caption{Intervention Recall by scenario (\%, higher is better, $\uparrow$).
Intervention-node counts give the denominators, including failed sessions.
Bold marks the best value in each column.}
\label{tab:scenario_results}

\begingroup
\fontsize{10}{12}\selectfont
\setlength{\tabcolsep}{3pt}
\setlength{\aboverulesep}{0pt}
\setlength{\belowrulesep}{0pt}
\setlength{\extrarowheight}{2pt}
\renewcommand{\arraystretch}{1.16}

\begin{tabularx}{\linewidth}{@{}l*{5}{>{\centering\arraybackslash}X}@{}}
\arrayrulecolor{black}\toprule[0.8pt]
\rowcolor{rawtitle}
\textbf{Model} &
\makecell{\textbf{Daily living}\\\textbf{\& services} $\uparrow$} &
\makecell{\textbf{Health \&}\\\textbf{caregiving} $\uparrow$} &
\makecell{\textbf{Travel \&}\\\textbf{mobility} $\uparrow$} &
\makecell{\textbf{Education}\\\textbf{\& work} $\uparrow$} &
\makecell{\textbf{Sports \&}\\\textbf{leisure} $\uparrow$} \\
\midrule
Intervention nodes & 144 & 146 & 77 & 82 & 106 \\
\midrule

\rowcolor{black!3}
MiniCPM-o 4.5
& 0.69 & 0.68 & 1.30 & 1.22 & 3.77 \\
MOSS-VL
& 9.72 & 4.79 & 3.90 & 7.32 & 1.89 \\

\rowcolor{black!3}
MOSS Preview
& 2.08 & 0.68 & 1.30 & 0.00 & 0.94 \\
Venus-Omni
& 3.47 & 0.68 & 3.90 & 2.44 & 0.00 \\

\rowcolor{black!3}
Gander
& 2.08 & 0.68 & 0.00 & 2.44 & 0.00 \\
AURA
& 17.36 & 13.01 & 10.39 & 3.66 & \textbf{14.15} \\

\rowcolor{black!3}
JoyAI
& \textbf{25.00} & \textbf{16.44} & \textbf{18.18}
& \textbf{29.27} & 12.26 \\
DuplexOmni (S1)
& 8.33 & 3.42 & 7.79 & 1.22 & 0.94 \\

\rowcolor{black!3}
Gemini 3.8 Live
& 6.25 & 1.37 & 2.60 & 2.44 & 3.77 \\
Qwen3.5 Plus
& 6.25 & 2.05 & 2.60 & 2.44 & 4.72 \\

\rowcolor{black!3}
Qwen3.5 Flash
& 6.94 & 2.05 & 2.60 & 4.88 & 3.77 \\
Qwen3.8 Flash
& 4.17 & 1.37 & 2.60 & 3.66 & 1.89 \\

\arrayrulecolor{black}\bottomrule[0.8pt]
\end{tabularx}

\endgroup
\end{table}

Table~\ref{tab:appendix_action} groups recall by the reference action. These four non-silence actions partition the 555 required nodes, while the wait action is evaluated through silence metrics. Correctness depends on the response content and timing, rather than agreement between a predicted behavior tag and the reference action. The answer and feedback groups have substantially fewer nodes than inform and warn, so individual matched events have a larger effect on their percentages.

\definecolor{rawtitle}{HTML}{E5F1EF}
\begin{table}[!t]
\centering
\caption{Intervention Recall (\%) by reference action.}
\label{tab:appendix_action}
\begingroup
\normalsize
\setlength{\tabcolsep}{3pt}
\renewcommand{\arraystretch}{1.22}
\begin{tabularx}{\linewidth}{@{}Xrrrr@{}}
\arrayrulecolor{black}\toprule[0.8pt]
\rowcolor{rawtitle}
\textbf{Model} & \shortstack{Inform\\362 nodes} & \shortstack{Warn\\149 nodes} & \shortstack{Feedback\\29 nodes} & \shortstack{Answer\\15 nodes} \\
\midrule
\rowcolor{black!3}
MiniCPM-o 4.5 & 1.66 & 0.67 & 0.00 & 6.67 \\
MOSS-VL & 5.25 & 8.05 & 0.00 & 6.67 \\
\rowcolor{black!3}
MOSS Preview & 1.66 & 0.00 & 0.00 & 0.00 \\
Venus-Omni & 1.93 & 2.01 & 0.00 & 6.67 \\
\rowcolor{black!3}
Gander & 1.10 & 0.00 & 0.00 & 13.33 \\
AURA & 11.88 & 15.44 & 6.90 & 13.33 \\
\rowcolor{black!3}
JoyAI & 18.23 & 26.17 & 17.24 & 6.67 \\
DuplexOmni (S1) & 3.59 & 6.71 & 3.45 & 6.67 \\
\rowcolor{black!3}
Gemini 3.8 Live & 3.04 & 1.34 & 0.00 & 40.00 \\
Qwen3.5 Plus & 3.59 & 3.36 & 0.00 & 20.00 \\
\rowcolor{black!3}
Qwen3.5 Flash & 4.97 & 1.34 & 0.00 & 20.00 \\
Qwen3.8 Flash & 1.93 & 2.01 & 0.00 & 33.33 \\
\arrayrulecolor{black}\bottomrule[0.8pt]
\end{tabularx}
\endgroup
\end{table}

\subsection{Preventive, Corrective, and Routine Assistance}
\label{app:intervention_functions}

We conducted a model-assisted descriptive classification of the 400 videos using task requests and reference annotations, without consulting evaluated-model outputs or scores. The unit was a non-wait reference node, identified by case and annotation index. Each of the 555 nodes received one primary function: preventive assistance averts an imminent error or adverse outcome; corrective or remedial assistance manages an existing deviation or harmful condition; routine task support provides requested information, counts, answers, or state reports. Counts are 68, 157, and 330, respectively. Case-level function sets are unions of node labels, so a video can contain multiple functions. The 110 silence-only videos retain empty function sets.

Classification follows the closest target problem rather than the action tag. An existing deviation takes precedence over a possible downstream consequence: stopping a child already touching an outlet is corrective even though it also aims to prevent electric shock. In \texttt{elder\_box\_m\_pos}, warning about a coffee table ahead is preventive, while post-fall guidance and seeking nearby help are remedial; both references use \texttt{[warn]}. For windows spanning preparation and execution, the primary label represents the best-supported service goal within the reference opportunity, without implying an invariant action phase or locating a violation's onset.

An initial model-assisted pass and internal cross-review covered all 400 cases and 555 nodes. Reviewers saw the initial labels, so annotation was not blinded or independent. Boundary adjudication also consulted source event descriptions and extracted video frames. Sixty-two nodes retain review flags, including a low-confidence leash reminder: the camera view does not establish leash absence, so the label follows the precautionary reminder rather than asserting an existing violation. A child reaching toward an armor exhibit is likewise classified as preventive because contact is not established and the subsequent event describes an adult stopping the reach. Technique-coaching nodes follow the task and reference's intended adjustment of current performance; this does not independently verify every technique judgment. The labels are not a human gold standard.

Temporal-state requirements were coded independently at task level. Seventy-four cases, including 69 intervention videos, require task-specific history such as cumulative counts, elapsed duration, ordered steps, or comparison with earlier observations; the other 326 are event-monitoring tasks under this definition. Generic monitoring instructions and multiple reference nodes alone do not establish the stateful label. In \texttt{cnt\_cafe\_cup}, the running count and final total require accumulated state, while quiet intervals still constrain when to speak.

Table~\ref{tab:intervention_function} groups the frozen main-experiment matches by function without rematching subsets or changing content judgments, response windows, or model outputs. Each function's denominator retains all its reference nodes, including those from failed sessions; previously credited partial outputs remain credited. Purely silent videos have no nodes in this table. Function-specific recall does not measure unwanted output, which must be read from the main silence and video-level metrics. Excluding the 62 flagged nodes as a boundary sensitivity check leaves 58 preventive, 122 corrective, and 313 routine nodes. MOSS-VL still covers corrective nodes more often than preventive nodes (8.20\% versus 1.72\%), while JoyAI retains higher preventive than routine recall (32.76\% versus 14.70\%). These comparisons describe the sampled tasks; task mix and observation conditions preclude interpreting them as causal effects of intervention function.

\subsection{Complete Instruction-Modality Ablation}
Table~\ref{tab:appendix_instruction_modality} reports the complete written-versus-spoken task-instruction results. The spoken condition replaces the written task with task TTS played once before the video. MiniCPM, Venus, and Qwen receive it through their audio input; MOSS-VL receives its transcript through the official SenseVoice path. Within each model, the general benchmark instruction, source media, model settings, and scoring rules remain fixed. All 400 videos per condition remain in the evaluation denominators, including failed sessions.

\begin{table}[!t]
\centering
\caption{Complete results for written and spoken task instructions on 400 videos per condition.}
\label{tab:appendix_instruction_modality}
\begingroup
\normalsize
\setlength{\tabcolsep}{3pt}
\renewcommand{\arraystretch}{1.15}
\begin{tabularx}{\linewidth}{@{}>{\raggedright\arraybackslash}Xrrrrr@{\hspace{8pt}}r@{}}
\arrayrulecolor{black}\toprule[0.8pt]
\multirow{2}{*}{\textbf{Task input}} & \multicolumn{2}{c}{Intervention (\%)} & \multirow{2}{*}{\shortstack{Full\\Correct (\%)}} & \multirow{2}{*}{\shortstack{Silence\\Success (\%)}} & \multirow{2}{*}{\shortstack{Unwanted\\/ min}} & \multirow{2}{*}{Failed} \\
 & Recall & F1 & & & & \\
\midrule
\rowcolor{ybtint}
\multicolumn{7}{l}{\textbf{MiniCPM-o 4.5}} \\*
Written & 1.44 & 2.63 & 1.38 & 95.45 & 0.14 & 0 \\
Spoken & 1.44 & 2.34 & 1.03 & 78.18 & 0.44 & 0 \\
\addlinespace[4pt]

\rowcolor{ybtint}
\multicolumn{7}{l}{\textbf{MOSS-VL}} \\*
Written & 5.77 & 7.06 & 4.48 & 82.73 & 1.41 & 0 \\
Spoken & 13.51 & 9.41 & 2.76 & 16.36 & 4.64 & 0 \\
\addlinespace[4pt]

\rowcolor{ybtint}
\multicolumn{7}{l}{\textbf{Venus-Omni}} \\*
Written & 1.98 & 3.18 & 2.07 & 77.27 & 0.48 & 0 \\
Spoken & 0.90 & 1.46 & 1.03 & 77.27 & 0.38 & 30 \\
\addlinespace[4pt]

\rowcolor{ybtint}
\multicolumn{7}{l}{\textbf{Qwen3.5 Flash}} \\*
Written & 4.14 & 5.39 & 3.10 & 80.91 & 1.34 & 31 \\
Spoken & 6.85 & 6.09 & 2.76 & 0.00 & 3.44 & 3 \\
\addlinespace[4pt]
\arrayrulecolor{black}\bottomrule[0.8pt]
\end{tabularx}
\endgroup
\end{table}

\section{Failure Cases Analysis}
\label{app:failure_cases}

We present four case studies that complement the failure
analysis in Section~\ref{sec:failure_taxonomy}.
They illustrate missed assistance, incorrect object
identification, unfulfilled task requirements, and unusable
output. The cases include three real-world videos and one
synthetic video. Source frames, reference windows, response
timelines, and descriptions of relevant outputs support
the analysis. Together, these materials connect each failure
to the observed situation and the assistance requested by
the user. These examples illustrate failures rather
than their frequency in the benchmark.

Green timeline segments mark required response windows,
red ticks substantive records, and gray ticks timestamp
or silence markers. Times are seconds from video onset,
with record onsets rounded to 0.001~s. Onsets indicate
first receipt by the harness, not the timing of every
later word. Reference intervals are half open and use
evaluation duration clipping. Tail records remain visible
but are excluded from reference-dependent judgments.
Failure types can overlap; each case highlights a specific
failure while retaining relevant timing and output details.

\subsection{Failing to correct an assembly error}
\label{app:failure_intervention}

\textbf{Model:} DuplexOmni (S1).
\textbf{Case:} \texttt{asm\_boom\_feedback\_01}.
\textbf{Source:} real-world subset.

\textbf{User task (English reference):}
Please watch me assemble the toy car, and correct me in
time if I assemble it incorrectly.

The sole response comments on the initial alignment at
0.724~s. No later output is recorded, including the
$[22,27)$ correction window. This is a missed intervention
after an earlier response, rather than a response that
arrives late. The initial comment does not satisfy
the later requirement to identify and correct the assembly
error when assistance is needed.
Figure~\ref{fig:appendix_failure_intervention} shows the
source frames and timeline, and
Table~\ref{tab:appendix_failure_intervention_reference}
lists the reference windows and required content.

\begingroup
\setlength{\intextsep}{6pt}
\begin{figure}[!t]

\centering

\begin{minipage}[t]{.485\linewidth}\centering
\includegraphics[width=\linewidth,height=3.4cm,keepaspectratio]{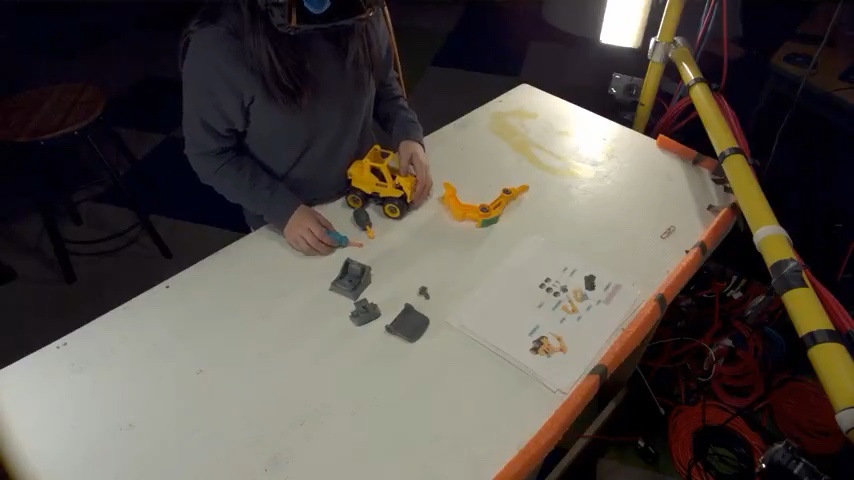}
\par\vspace{3pt}\textbf{1 s}: Assembly before the error
\end{minipage}\hfill
\begin{minipage}[t]{.485\linewidth}\centering
\includegraphics[width=\linewidth,height=3.4cm,keepaspectratio]{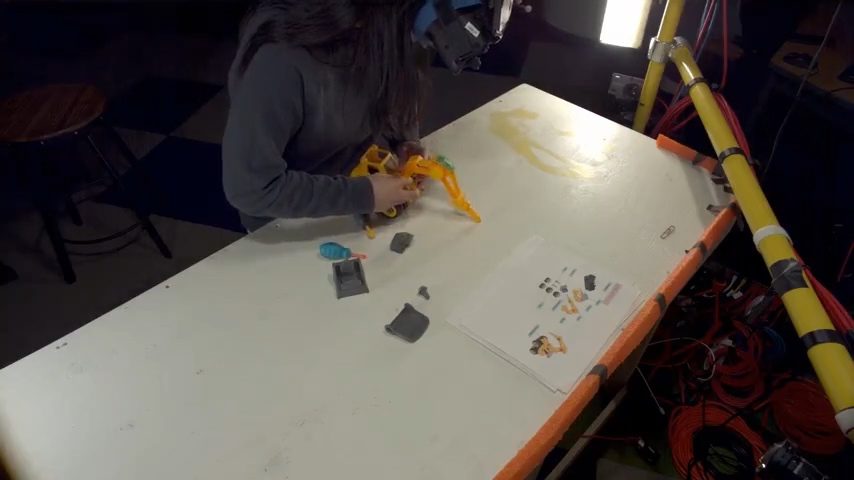}
\par\vspace{3pt}\textbf{24 s}: Reversed arm in the correction window
\end{minipage}

\par\vspace{5pt}

\begin{tikzpicture}[x=1cm,y=1cm,font=\normalsize]
\node[anchor=west] at (0,1.08) {Reference};
\node[anchor=west] at (0,.48) {Records};
\draw[ybgrey] (1.8,1.08) -- (14.3,1.08);
\draw[ybgrey] (1.8,.48) -- (14.3,.48);
\fill[ybrule] (6.38333,.97) rectangle (7.42500,1.19);
\draw[ybneg,line width=1pt] (1.95084,.34) -- (1.95084,.62);
\draw[black] (1.80000,.11) -- (1.80000,.19);
\node[anchor=north] at (1.80000,.08) {0 s};
\draw[black] (4.92500,.11) -- (4.92500,.19);
\node[anchor=north] at (4.92500,.08) {15 s};
\draw[black] (8.05000,.11) -- (8.05000,.19);
\node[anchor=north] at (8.05000,.08) {30 s};
\draw[black] (11.17500,.11) -- (11.17500,.19);
\node[anchor=north] at (11.17500,.08) {45 s};
\draw[black] (14.30000,.11) -- (14.30000,.19);
\node[anchor=north] at (14.30000,.08) {60 s};
\end{tikzpicture}

\caption{Source frames and session timeline: DuplexOmni (S1) on \texttt{asm\_boom\_feedback\_01}.}

\label{fig:appendix_failure_intervention}

\end{figure}

\begin{table}[!t]
\centering
\caption{Reference windows and required content for \texttt{asm\_boom\_feedback\_01}.}
\label{tab:appendix_failure_intervention_reference}
\normalsize
\renewcommand{\arraystretch}{1.15}
\setlength{\tabcolsep}{5pt}
\begin{tabularx}{\linewidth}{@{}p{2.4cm}p{1.65cm}X@{}}
\toprule
\textbf{Window (s)} & \textbf{Action} & \textbf{Required content (English reference)} \\
\midrule
$[0,22)$ & wait & Remain silent. \\
$[22,27)$ & feedback & The excavator arm is on backwards; it should bend toward the rear. \\
$[27,60)$ & wait & Remain silent. \\
\bottomrule
\end{tabularx}
\end{table}
\endgroup

\subsection{Mistaking a thermos for a guitar}
\label{app:failure_grounding}

\textbf{Model:} AURA.
\textbf{Case:} \texttt{orbit\_mug\_06}.
\textbf{Source:} real-world subset.

\textbf{User task (English reference):}
Please identify the object in my hand and tell me what it is.

The substantive response begins at 5.075~s, inside the
$[1.5,5.5)$ reference window, but identifies a black electric
guitar instead of the stainless steel thermos. Although
the response begins within the required interval, its
central object identification is incorrect.
The remaining nine records are timestamp strings.
The qualitative grounding error concerns the stated
identity, independent of these separate output artifacts.
Figure~\ref{fig:appendix_failure_grounding} shows the source
frames and timeline;
Table~\ref{tab:appendix_failure_grounding_reference}
provides the references for evaluating the requested identification.

\input{floats/appendix-failure-grounding}

\subsection{Saying guests were greeted instead of greeting them}
\label{app:failure_fulfillment}

\textbf{Model:} MOSS-VL.
\textbf{Case:} \texttt{robot\_host\_rest}.
\textbf{Source:} synthetic subset.

\textbf{User task (English reference):}
Please watch for guests entering the restaurant, greet
them when they arrive, and tell them they may take a seat.

The only response starts at 7.189~s, within the $[6,11)$
greeting window. It reports that the guests have already
been greeted, but never addresses them with a greeting
or an invitation to sit. Describing the requested action
as completed does not perform the guest-facing assistance
specified in the task. This case illustrates a
task-fulfillment failure: the response is timely and
relevant to the event, but does not fulfill the requested
assistance. Figure~\ref{fig:appendix_failure_fulfillment}
shows the source frames and timeline, and
Table~\ref{tab:appendix_failure_fulfillment_reference}
provides the references.

\begingroup
\setlength{\intextsep}{6pt}
\begin{figure}[!t]

\centering

\begin{minipage}[t]{.485\linewidth}\centering
\includegraphics[width=\linewidth,height=3.4cm,keepaspectratio]{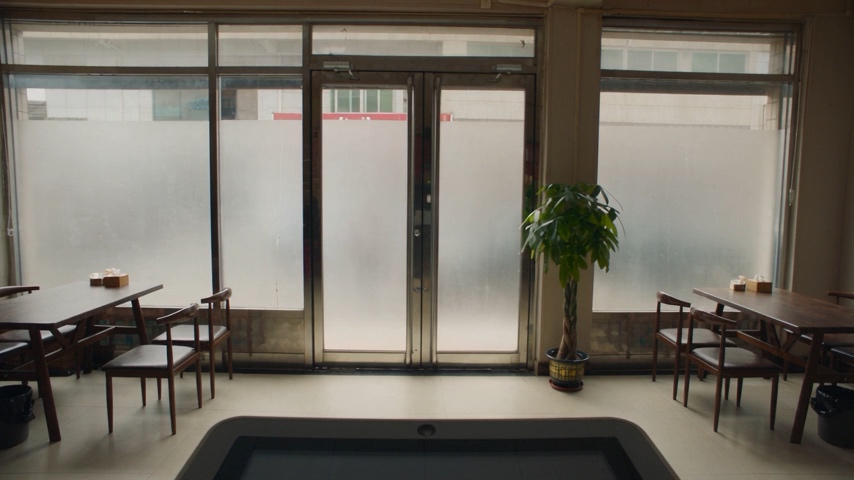}
\par\vspace{3pt}\textbf{3 s}: Entrance before the guests arrive
\end{minipage}\hfill
\begin{minipage}[t]{.485\linewidth}\centering
\includegraphics[width=\linewidth,height=3.4cm,keepaspectratio]{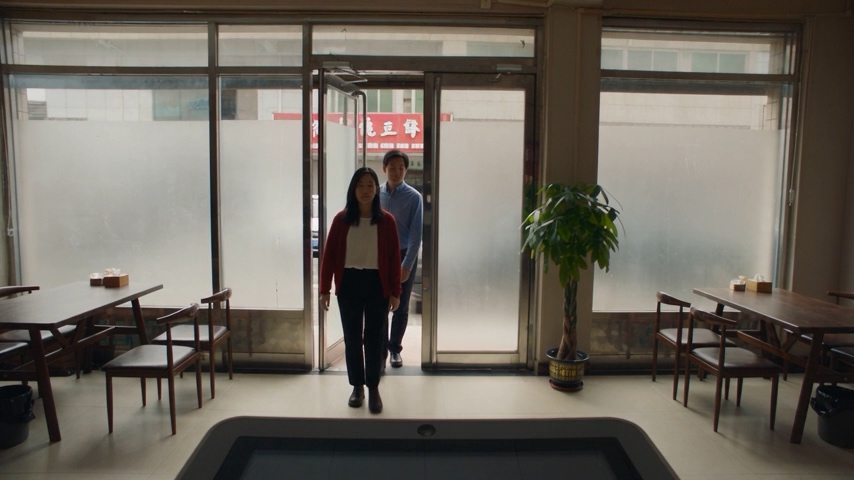}
\par\vspace{3pt}\textbf{9 s}: Two guests enter the restaurant
\end{minipage}

\par\vspace{5pt}

\begin{tikzpicture}[x=1cm,y=1cm,font=\normalsize]
\node[anchor=west] at (0,1.08) {Reference};
\node[anchor=west] at (0,.48) {Records};
\draw[ybgrey] (1.8,1.08) -- (14.3,1.08);
\draw[ybgrey] (1.8,.48) -- (14.3,.48);
\fill[ybrule] (4.25902,.97) rectangle (6.30820,1.19);
\draw[ybneg,line width=1pt] (4.74637,.34) -- (4.74637,.62);
\draw[black] (1.80000,.11) -- (1.80000,.19);
\node[anchor=north] at (1.80000,.08) {0 s};
\draw[black] (5.89836,.11) -- (5.89836,.19);
\node[anchor=north] at (5.89836,.08) {10 s};
\draw[black] (9.99672,.11) -- (9.99672,.19);
\node[anchor=north] at (9.99672,.08) {20 s};
\draw[black] (14.09508,.11) -- (14.09508,.19);
\node[anchor=north] at (14.09508,.08) {30 s};
\end{tikzpicture}

\caption{Source frames and session timeline: MOSS-VL on \texttt{robot\_host\_rest}.}

\label{fig:appendix_failure_fulfillment}

\end{figure}

\begin{table}[!t]
\centering
\caption{Reference windows and required content for \texttt{robot\_host\_rest}.}
\label{tab:appendix_failure_fulfillment_reference}
\normalsize
\renewcommand{\arraystretch}{1.15}
\setlength{\tabcolsep}{5pt}
\begin{tabularx}{\linewidth}{@{}p{2.4cm}p{1.65cm}X@{}}
\toprule
\textbf{Window (s)} & \textbf{Action} & \textbf{Required content (English reference)} \\
\midrule
$[0,6)$ & wait & Remain silent. \\
$[6,11)$ & inform & Welcome, please feel free to take a seat. \\
$[11,30.08)$ & wait & Remain silent. \\
\bottomrule
\end{tabularx}
\end{table}
\endgroup

\subsection{Outputting timestamps instead of reporting a leak in time}
\label{app:failure_unusable}

\textbf{Model:} AURA.
\textbf{Case:} \texttt{roof\_leak}.
\textbf{Source:} real-world subset.

\textbf{User task (English reference):}
Please monitor the room and tell me the location and
situation if you observe a leak or persistent dripping.

The record at 3.066~s is only
\texttt{\textless{}2.5 seconds\textgreater{}}, inside the
$[2,5)$ leak-report window. The second $[22,25)$ reference
window likewise contains only a timestamp record at
23.200~s. The first substantive statement arrives at
57.093~s. Timestamp strings do not convey the requested
location or situation, so the presence of a record within
a response window does not itself constitute usable
assistance. These are recorded interface outputs; this
evidence does not establish that the strings were spoken
aloud. The record after the reference horizon is shown
for completeness and is not used for reference-dependent
classification.
Figure~\ref{fig:appendix_failure_unusable} shows the source
frames and timeline, and
Table~\ref{tab:appendix_failure_unusable_reference}
lists the reference windows and required content.

\input{floats/appendix-failure-unusable}
\section{Cases Across Users, Scenarios, and Tasks}
\label{app:case_gallery}
\begingroup
\captionsetup{hypcap=false}
\captionsetup[figure]{skip=4pt}
\captionsetup[table]{skip=4pt}

We present 14 failure cases across user groups, five scenarios,
and five failure categories. Eight videos are real and six
are controlled synthetic videos. These cases illustrate
failure mechanisms rather than their benchmark frequency.
Together, these examples show how assistance can fail through
inappropriate timing, inaccurate interpretation of the situation,
or incomplete fulfillment of the user's specific requested task.
The paired cases also distinguish between producing a response,
providing the information the task requires, and intervening only
when the reference explicitly calls for assistance.
The analysis considers both what the assistant says and whether
that response satisfies the user's need at the relevant moment.
Table~\ref{tab:case_gallery_coverage} summarizes coverage;
Figures~\ref{fig:case_gallery_1}--\ref{fig:case_gallery_7}
pair video evidence with response analysis.

Source frames appear on the left; reference requirements,
model responses, and analysis appear on the right.
Frames within each panel follow chronological order.
This arrangement allows readers to compare the observed evidence,
expected behavior, and recorded output while checking whether each
response occurs within the required temporal window.
Green timeline segments mark response windows, gray segments
mark required silence, and red points mark recorded response
onsets. Responses are translated into English; ellipses mark
omitted text, and protocol tags are omitted for readability.
Frame aspect ratios are preserved; C07 includes an enlarged
view of the original image pixels.
Interpretation combines the displayed frames with annotations
and recorded outputs, since individual images cannot independently
establish every event or the content of an auditory signal.
Case-specific discussion below explains these distinctions while
retaining the evidence limitations associated with visual
observations, annotated events, and interface records.

\par\medskip
\noindent\begin{minipage}{\linewidth}
\centering
\small
\setlength{\tabcolsep}{4pt}
\renewcommand{\arraystretch}{1.08}
\captionof{table}{Coverage of the 14 cases. Population labels
 describe benchmark tasks.}
\label{tab:case_gallery_coverage}
\begin{tabularx}{\linewidth}{@{}l
>{\raggedright\arraybackslash}p{0.26\linewidth}
>{\raggedright\arraybackslash}X c c@{}}
\toprule
Case & Service population & User need & Scene & Failure \\
\midrule
C01 & Adults & Summarize after coffee preparation & S1 & F3 \\
C02 & Infants and toddlers & Choking alert & S2 & F2 \\
C03 & Older adults & Timely help & S2 & F1 \\
C04 & Cognitive impairment & Remember item location & S1 & F4 \\
C05 & Multi-person settings & Fall alert & S2 & F2 \\
C06 & Visual impairment & Hazard warning & S3 & F2 \\
C07 & Color-vision deficiency & Wire-connection guidance & S4 & F2 \\
C08 & Mobility impairment & Movement progress & S3 & F3 \\
C09 & Schoolchildren & Clarify a question & S4 & F4 \\
C10 & Multi-person settings & Cross-player turn order & S5 & F1 \\
C11 & Pets & Pet-care need & S2 & F2 \\
C12 & Hearing impairment & Sound amid background talk & S1 & F4 \\
C13 & Schoolchildren & Do not interrupt & S4 & F1 \\
C14 & Hearing impairment & Usable alarm notification & S1 & F5 \\
\bottomrule
\end{tabularx}

\end{minipage}

\smallskip
\noindent\textbf{Scenarios.}
S1: daily living and services; S2: health and caregiving;
S3: travel and mobility; S4: education and work;
S5: sports and leisure.

\noindent\textbf{Failure categories.}
F1: intervention control; F2: situational grounding;
F3: temporal tracking; F4: task fulfillment;
F5: unusable output.

\par\addvspace{9pt}
\noindent\begin{minipage}{\linewidth}
\begingroup
\centering
\includegraphics[width=\linewidth]{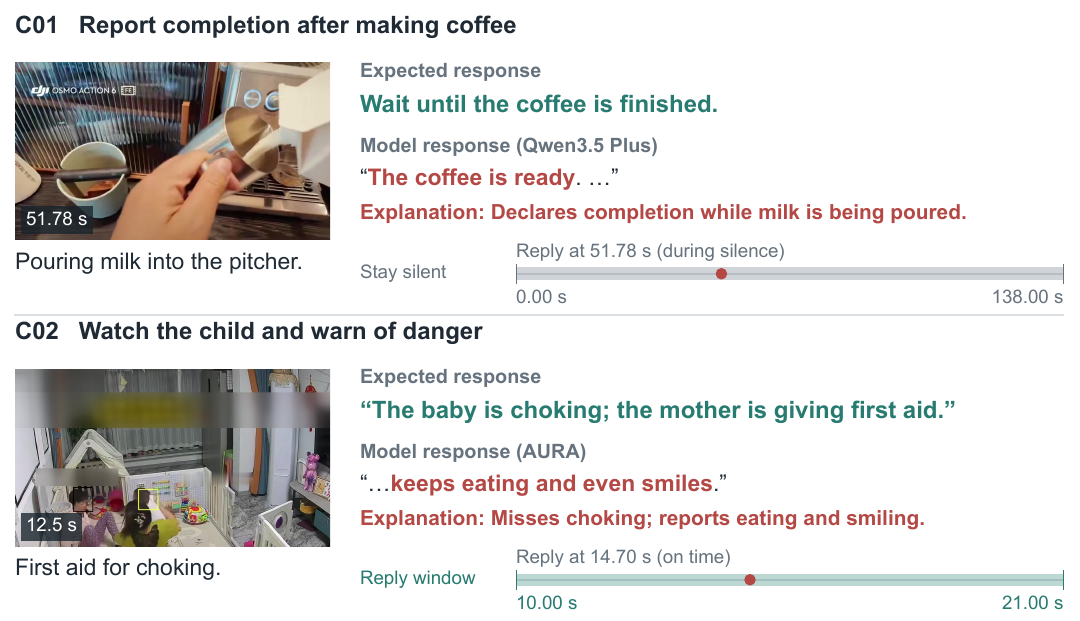}\par
\captionof{figure}{C01 (\texttt{coffee\_ego}; real video) and C02 (\texttt{baby\_danger}; real video).}
\label{fig:case_gallery_1}
\endgroup
\par\smallskip
\noindent Figure~\ref{fig:case_gallery_1} contrasts temporal tracking and situational grounding failures.
C01 declares completion while milk is still being poured,
confusing an intermediate step with the requested endpoint.
C02 responds on time but describes eating and smiling instead
of the annotated choking event. The choking interpretation
relies on video annotations; still frames alone cannot establish it.
\end{minipage}

\par\addvspace{9pt}
\noindent\begin{minipage}{\linewidth}
\begingroup
\centering
\includegraphics[width=\linewidth]{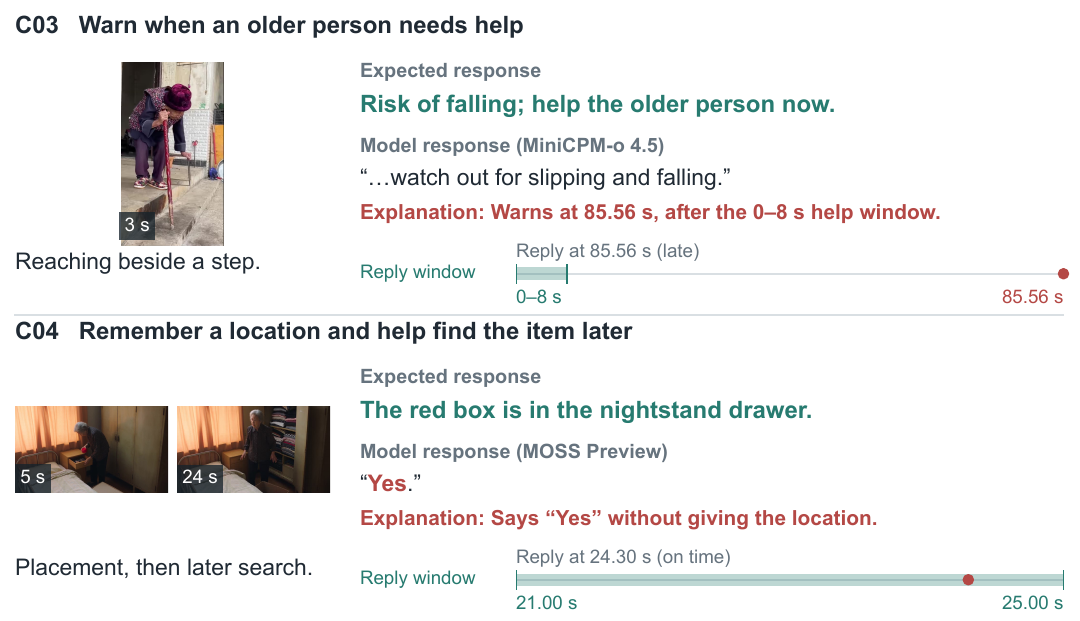}\par
\captionof{figure}{C03 (\texttt{elder\_mobility}; real video) and C04 (\texttt{mon\_alz\_find\_item}; controlled synthetic).}
\label{fig:case_gallery_2}
\endgroup
\par\smallskip
\noindent Figure~\ref{fig:case_gallery_2} illustrates late assistance and incomplete recall.
C03 warns after the required help window, missing the
opportunity for timely intervention. C04 responds within
its window but offers only an acknowledgment, omitting
the remembered item location needed to fulfill the request
for help finding the item.
\end{minipage}

\par\addvspace{9pt}
\noindent\begin{minipage}{\linewidth}
\begingroup
\centering
\includegraphics[width=\linewidth]{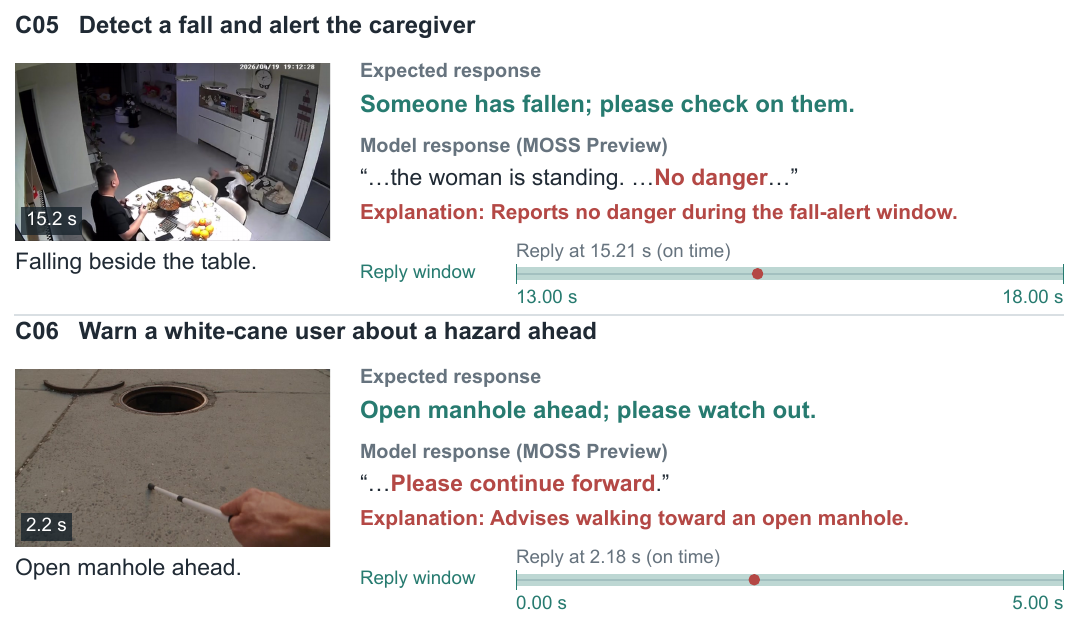}\par
\captionof{figure}{C05 (\texttt{dining\_help}; real video) and C06 (\texttt{blind\_cane\_hole}; controlled synthetic).}
\label{fig:case_gallery_3}
\endgroup
\par\smallskip
\noindent Figure~\ref{fig:case_gallery_3} presents two timely responses with incorrect hazard assessments.
C05 reports no danger during a fall, while C06
advises continued movement toward an open manhole.
The failure concerns the content of the guidance despite
its onset falling within the required window.
\end{minipage}

\par\addvspace{9pt}
\noindent\begin{minipage}{\linewidth}
\begingroup
\centering
\includegraphics[width=\linewidth]{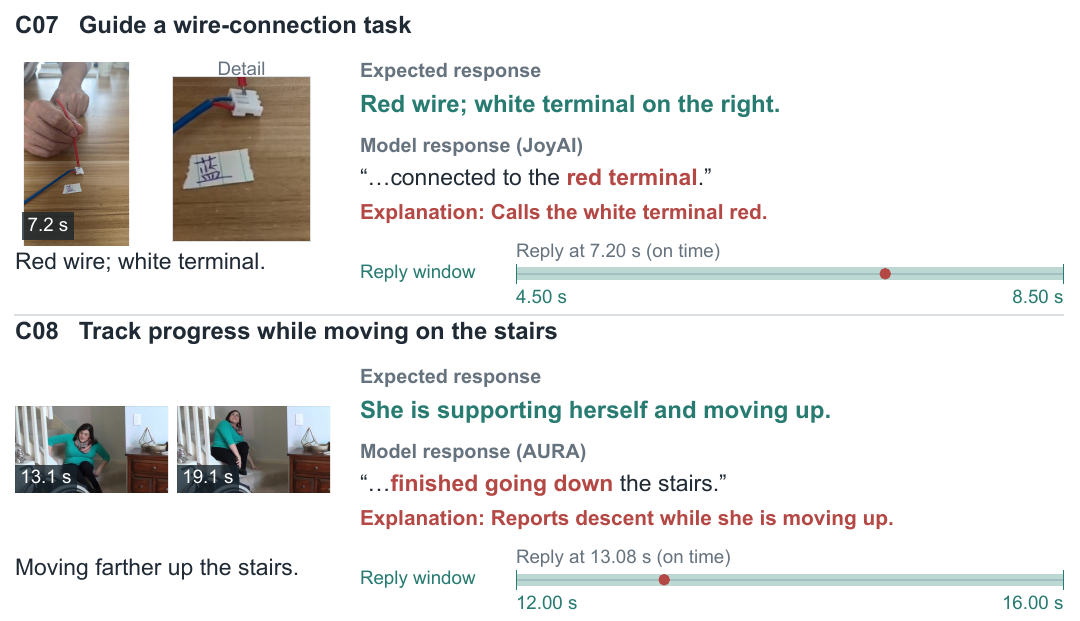}\par
\captionof{figure}{C07 (\texttt{cblind\_wire\_01}; controlled synthetic) and C08 (\texttt{wheel\_stairs\_paralyzed}; real video).}
\label{fig:case_gallery_4}
\endgroup
\par\smallskip
\noindent Figure~\ref{fig:case_gallery_4} highlights errors in attribute identification and movement tracking.
C07 calls the white terminal red, undermining the requested
wire-connection guidance. C08 reports completed descent
while the person is moving upward. Both responses begin
within their reference windows but misrepresent visual
details essential to the task.
\end{minipage}

\par\addvspace{9pt}
\noindent\begin{minipage}{\linewidth}
\begingroup
\centering
\includegraphics[width=\linewidth]{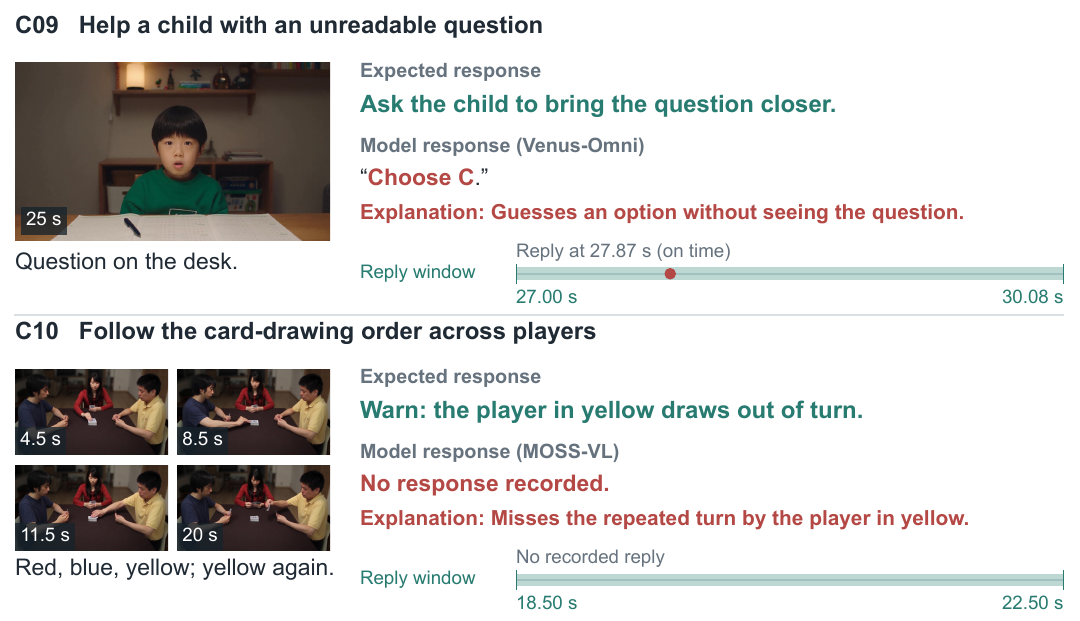}\par
\captionof{figure}{C09 (\texttt{tutor\_hold}; controlled synthetic) and C10 (\texttt{life\_poker\_turns}; controlled synthetic).}
\label{fig:case_gallery_5}
\endgroup
\par\smallskip
\noindent Figure~\ref{fig:case_gallery_5} pairs an unsupported answer with a missed intervention.
C09 selects an option instead of requesting a clearer view
of the question. The evidence establishes insufficient
support for the answer, not that the option is wrong.
C10 has no recorded response during the intervention window,
leaving the repeated turn by the player in yellow unaddressed.
\end{minipage}

\par\addvspace{9pt}
\noindent\begin{minipage}{\linewidth}
\begingroup
\centering
\includegraphics[width=\linewidth,height=0.34\textheight,keepaspectratio]{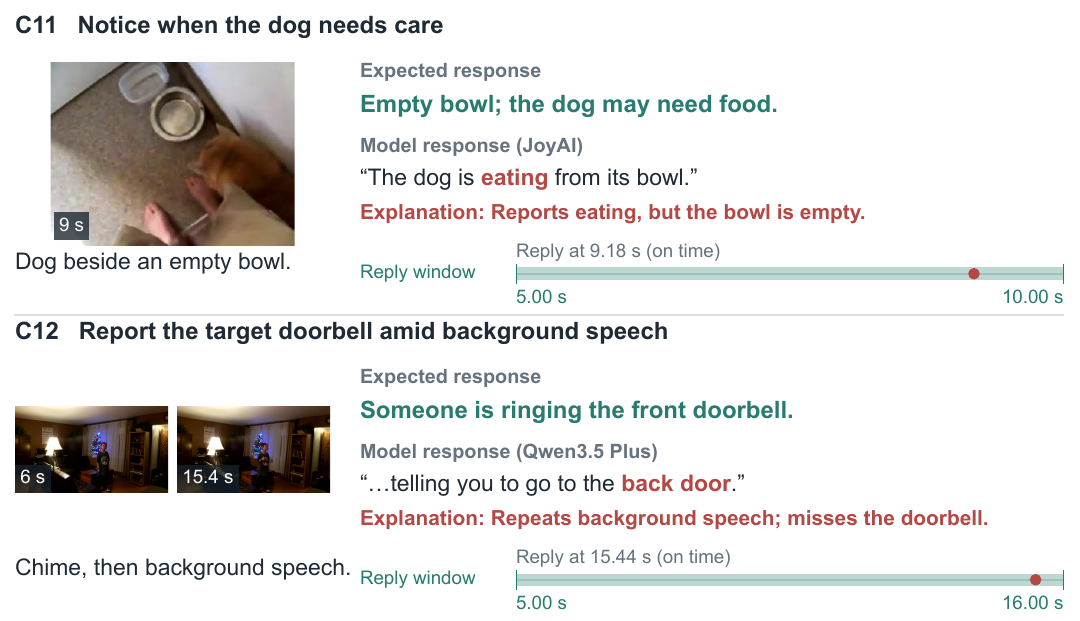}\par
\captionof{figure}{C11 (\texttt{pet\_bowl\_ego}; real video) and C12 (\texttt{deaf\_sign\_home}; real video).}
\label{fig:case_gallery_6}
\endgroup
\par\smallskip
\noindent Figure~\ref{fig:case_gallery_6} illustrates missed care cues and difficulty separating target sounds from background speech.
C11 describes eating despite the empty bowl, missing the
cue relevant to the dog's care needs. C12 repeats background
instructions instead of reporting the front-door chime
as required by the task.
The chime is identified in video annotations, and speech
transcripts confirm the competing background instructions.
\end{minipage}

\par\addvspace{9pt}
\noindent\begin{minipage}{\linewidth}
\begingroup
\centering
\includegraphics[width=\linewidth,height=0.34\textheight,keepaspectratio]{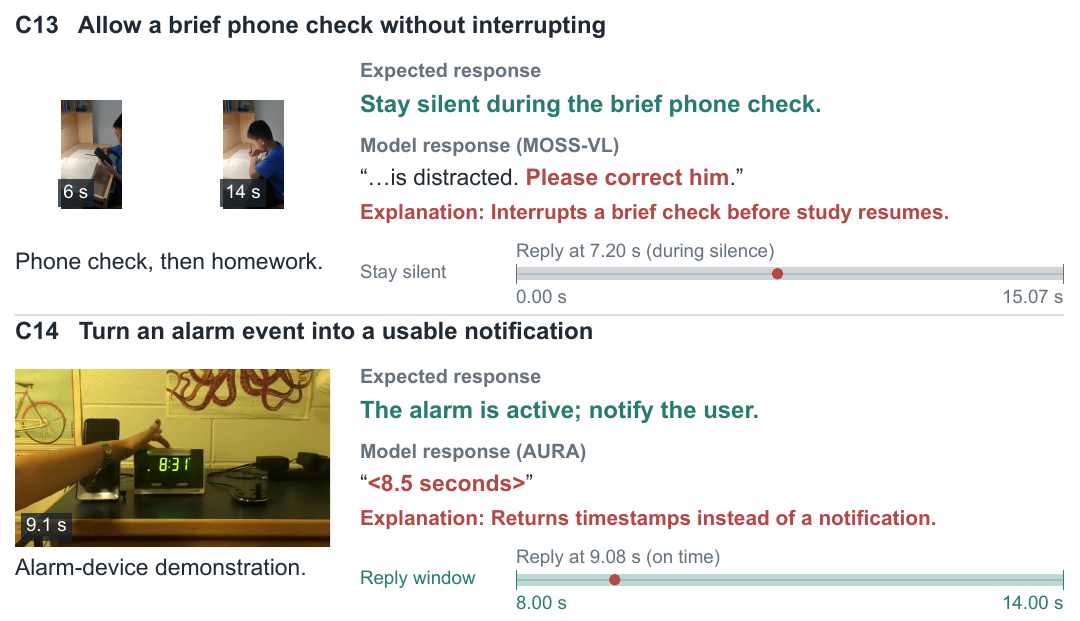}\par
\captionof{figure}{C13 (\texttt{kid\_phone\_neg}; controlled synthetic) and C14 (\texttt{deaf\_wake}; real video).}
\label{fig:case_gallery_7}
\endgroup
\par\smallskip
\noindent Figure~\ref{fig:case_gallery_7} contrasts an unnecessary intervention with an unusable output.
C13 interrupts a permitted brief phone check during a
required silence interval, despite the subsequent return
to homework. C14 returns a timestamp within the response
window without an informative alarm notification.
This is a recorded interface output; the evidence does
not establish that the timestamp was spoken aloud.
\end{minipage}

\par
\endgroup

\end{document}